\documentclass[Afour,sageh,times]{sagej}

\usepackage{times}
\usepackage{color, colortbl}
\usepackage[dvipsnames]{xcolor}
\usepackage{multicol}
\usepackage{amsfonts}
\usepackage{amsmath}
\usepackage{dsfont}
\usepackage{subfigure}
\usepackage{authblk}
\usepackage{cases}
\usepackage{xspace}
\usepackage[acronyms, shortcuts]{glossaries}
\usepackage{siunitx}
\usepackage[textfont=md,font=footnotesize]{caption}

\usepackage{moreverb,url}

\usepackage[colorlinks,bookmarksopen,bookmarksnumbered,citecolor=red,urlcolor=red]{hyperref}
\usepackage[capitalize,noabbrev,nameinlink]{cleveref}

\usepackage{xcolor}
\newcommand{\revised}[1]{{\leavevmode#1}}

\glsdisablehyper

\newacronym{ioc}{IOC}{inverse optimal control}
\newacronym{mpc}{MPC}{model predictive control}
\newacronym{lqr}{LQR}{linear-quadratic regulator}
\newacronym{lq}{LQ}{linear-quadratic}
\newacronym{kkt}{KKT}{Karush–Kuhn–Tucker}
\newacronym{irl}{IRL}{inverse reinforcement learning}
\newacronym{mle}{MLE}{maximum likelihood estimation}
\newacronym[longplural=open-loop Nash equilibria,plural=OLNE]{olne}{OLNE}{open-loop Nash equilibrium}
\newacronym[longplural={partially observable Markov decision processes}]{pomdp}{POMDP}{partially observable Markov decision process}
\newacronym{svo}{SVO}{social value orientation}
\newacronym{ukf}{UKF}{unscented Kalman filter}
\newacronym{ibr}{IBR}{iterated best response}
\newacronym{awgn}{AWGN}{additive white Gaussian noise}
\newacronym{iqr}{IQR}{interquartile range}

\newenvironment{rcases}
  {\left.\begin{aligned}}
  {\end{aligned}\right\rbrace}
\newcommand{\ie}{i.e.\@\xspace}
\newcommand{\eg}{e.g.\@\xspace}

\newcommand{\norm}[1]{\left\lVert#1\right\rVert}
\newcommand{\given}{\mid}
\newcommand{\mbb}{\mathbb}
\newcommand{\mc}{\mathcal}
\newcommand{\mbf}{\mathbf}
\newcommand{\numplayers}{N}
\newcommand{\player}[1]{\text{Player-$#1$}}
\newcommand{\dynamics}[1]{f_{#1}}
\newcommand{\capdynamics}{\mbf{F}}
\newcommand{\state}{x}
\newcommand{\control}[1]{u^{#1}}
\newcommand{\bstate}{\mathbf{x}}
\newcommand{\bcontrol}[1]{\mathbf{u}^{#1}}
\newcommand{\bcontrolt}{u_t}
\newcommand{\xdim}{n}
\newcommand{\udim}[1]{m^{#1}}
\newcommand{\horizon}{T}
\newcommand{\obshorizon}{{\tilde{T}}}
\newcommand{\cost}[1]{J^{#1}}
\newcommand{\runningcost}[1]{g^{#1}}
\newcommand{\costparamfloor}{c}
\newcommand{\costate}[1]{\lambda^{#1}}
\newcommand{\bcostate}[1]{\boldsymbol{\lambda}^{#1}}
\newcommand{\costparams}[1]{\theta^{#1}}
\newcommand{\costparamdim}[1]{k^{#1}}

\newcommand{\observation}{y}
\newcommand{\bobservation}{\mathbf{\observation}}
\newcommand{\noise}{n}
\newcommand{\noisecov}{\Sigma}
\newcommand{\game}{\Gamma}
\newcommand{\kktresidual}{\mathbf{G}}
\newcommand{\indicator}{\ensuremath{\mathbf{1}}\xspace}
\newcommand{\obsmap}{h}
\newcommand{\position}{p}
\newcommand{\speed}{v}
\newcommand{\heading}{\psi}
\newcommand{\acceleration}{a}
\newcommand{\yawrate}{\omega}
\newcommand{\weight}{w}

\newcommand{\goalmap}{\textnormal{d}}
\newcommand{\strategy}{\gamma}
\newcommand{\infoset}{\mc{I}}

\newcommand{\remove}[1]%
    {\textcolor{red}{#1}}

\newcommand{\example}[1]%
{
\vspace{0.15cm}
\noindent \textit{\textbf{Running example:} #1}
\vspace{0.15cm}
}

\newtheorem{remark}{Remark}
\newtheorem{assumption}{Assumption}
\newtheorem{definition}{Definition}

\graphicspath{{./figs/}}
\DeclareGraphicsExtensions{.pdf,.png}

\newcommand\BibTeX{{\rmfamily B\kern-.05em \textsc{i\kern-.025em b}\kern-.08em
T\kern-.1667em\lower.7ex\hbox{E}\kern-.125emX}}
\newcommand{\vegascale}{0.5}

\def\volumeyear{2018}

\begin{document}

\runninghead{Peters et. al.}

\title{Online and Offline Learning of Player Objectives from Partial Observations \\in Dynamic Games}

\author[1,3]{Lasse Peters}
\author[2]{Vicen\c{c} Rubies-Royo}
\author[2]{Claire J. Tomlin}
\author[1]{Laura Ferranti}
\author[1]{Javier Alonso-Mora}
\author[3]{Cyrill~Stachniss}
\author[4]{David Fridovich-Keil}
\affil[1]{Delft University of Technology}
\affil[2]{University of California, Berkeley}
\affil[3]{University of Bonn}
\affil[4]{The University of Texas at Austin}

\corrauth{Lasse Peters,
Delft University of Technology,
Delft, The Netherlands.}

\email{l.peters@tudelft.nl}

\begin{abstract}
Robots deployed to the real world must be able to interact with other agents in their environment.
Dynamic game theory provides a powerful mathematical framework for modeling scenarios in which agents have individual objectives and interactions evolve over time.
However, a key limitation of such techniques is that they require a-priori knowledge of all players' objectives.
In this work, we address this issue by proposing a novel method for learning players' objectives in continuous dynamic games from noise-corrupted, partial state observations.
Our approach learns objectives by coupling the estimation of unknown cost parameters of each player with inference of unobserved states and inputs through Nash equilibrium constraints.
By coupling past state estimates with future state predictions, our approach is amenable to simultaneous online learning and prediction in receding horizon fashion.
We demonstrate our method in several simulated traffic scenarios in which we recover players' preferences for, e.g., desired travel speed and collision-avoidance behavior.
Results show that our method reliably estimates game-theoretic models from noise-corrupted data that closely matches ground-truth objectives, consistently outperforming state-of-the-art approaches.
\end{abstract}


\maketitle

\section{Introduction}
\label{sec:intro}

\begin{figure}
  \centering
  \includegraphics[scale=0.44]{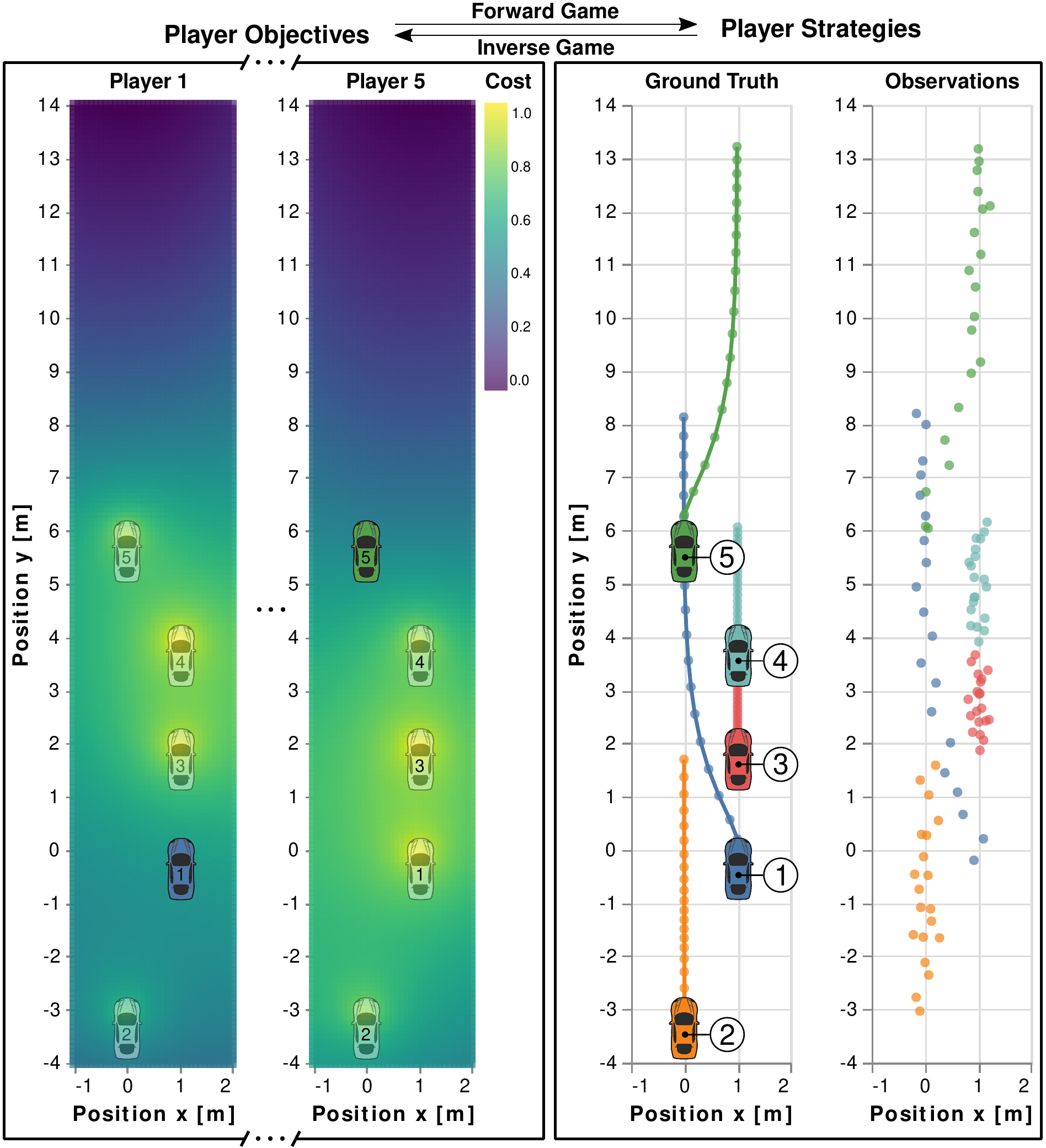}
  \caption{5-player highway driving scenario, modeled as a dynamic game. Solving
    the ``forward'' problem amounts to finding optimal trajectories (right) for
    all cars, given their objectives (left). In contrast, this paper addresses
    the ``inverse,''~i.e., it estimates the objectives of each player given
    noise-corrupted observations of each agent's trajectories. For example, our method can
    infer properties such as the degree to which each player wishes to keep a
    safe distance from others (heatmap, left). These learned objectives
    constitute an abstract model which can be used to predict players' actions
    in the future.}
  \label{fig:frontfig}
\end{figure}

To operate safely and efficiently in environments shared with other agents, robots must be able to predict the effects of their actions on the decisions of others.
In many such settings, agents do not form a single team that shares a joint objective.
Instead, each agent may have an individual objective, encoded by a cost function which they optimize unilaterally.
Unless the objectives of all agents are perfectly aligned, agents must therefore compete to minimize their own cost, while accounting for the strategic behavior of others.
For example, consider the highway navigation scenario in \cref{fig:frontfig}.
Here, each driver travels along the highway with an individual objective that encodes their preferences for speed, acceleration, and proximity to other cars.
In heavy traffic, the objectives of drivers may conflict.
For instance, if car 1 (blue) wishes to maintain its speed, it must overtake the slower vehicles in front.
At the same time, however, the faster car 2 (orange) may wish maintain its speed and but would be forced to decelerate if the driver of car 1 changes lanes.

Mathematically, such interactions of multiple agents with individual, potentially conflicting objectives are characterized by a \emph{noncooperative dynamic game}.
The theory underpinning dynamic games is well established~\citep{isaacs1954differential,basar1998gametheorybook} and recent work has put forth efficient algorithms to determine equilibrium solutions to these problems, given players' objectives~\citep{fridovich2020icra,di2019cdc}.
The \emph{forward} game problem is depicted in~\cref{fig:frontfig} (left to right) for the highway driving scenario: given the cost functions of all players (left), a forward game solver computes their rational strategies and corresponding future trajectories (right).

Unfortunately, the objectives of agents in a scene are often not known a priori.
Therefore, in order for game-theoretic methods to find practical application in fields such as robotics, it is imperative to recover these objectives from data.
This \emph{inverse} dynamic game problem is illustrated in~\cref{fig:frontfig} (right to left) for the highway driving scenario: given observations of player's strategies (right), an inverse game solver recovers objectives (left) which explain the observed behavior.
This inverse dynamic game problem is the focus of this work.

The challenge of recovering objectives from observed behavior has been extensively studied in the literature on \ac{ioc} \citep{kalman1964jbe,mombaur2010ar,albrecht2011humanoids} and \ac{irl} \citep{ng2000icml,ziebart2008aaai}.
\revised{Unfortunately, however, these methods are fundamentally limited to the single-player setting.}
\revised{While recent efforts extend these ideas to multi-agent \ac{irl}~\citep{vsovsic2016inverse,natarajan2010multi}, those approaches are limited to games with \emph{potential} cost structures~\citep{monderer1996potential} and do not directly apply in general noncooperative settings.}
While initial work extends \ac{ioc} methods to address this limitation \citep{rothfuss2017ifac,inga2019arxiv,awasthi2020acc}, these inverse dynamic game solvers rely upon full observation of states and inputs of all players.

The main contribution of this work is a novel method for learning  player's objectives in noncooperative dynamic games from only noise-corrupted, partial state observations.
In addition to learning a cost model for all players, our method also recovers a forward game solution consistent with the learned objectives by enforcing equilibrium constraints on latent trajectory estimates.
This bilevel formulation further allows to couple observed and predicted behavior to recover player's objectives even from temporally-incomplete interactions.
As a result, our approach is amenable for online learning and prediction in receding horizon fashion.

This paper builds upon and extends our earlier work~\citep{peters2021rss}.
In this work, we provide a more in-depth analysis of that approach.
Additionally, while our original work was limited to offline operation and could therefore only recover players' objectives for interactions which had already occurred, in this work we remove this requirement.

We evaluate our method in extensive Monte Carlo simulations in several traffic scenarios with varying numbers of players and interaction geometries. %
Empirical results show that our approach is more robust to partial state observations, measurement noise, and unobserved time-steps than existing methods, and consequently is more suitable for predicting agents' actions in the future.

\section[Prior Work]{Prior Work}%
\label{sec:related}

We begin by discussing recent advances in the well-studied area of \ac{ioc}.
While methods from that field address only single-player, cooperative settings, this body of work exposes many of the important mathematical and algorithmic concepts that appear in games.
We discuss how some of these approaches have been applied in the noncooperative multi-player setting and emphasize the connections between existing approaches and our contributions.

\subsection{Single-Player Inverse Optimal Control}\label{sec:related-single-player}

The \ac{ioc} problem has been extensively studied since the well-known work of \citet{kalman1964jbe}.
In the context of \ac{irl}, early formulations such as that of \citet{ng2000icml} and maximum-entropy variants~\citep{ziebart2008aaai,kretzschmar2016socially} have proven successful in treating problems with discrete state and control sets.
In robotic applications, optimal control problems typically involve decision variables in a continuous domain.
Hence, recent work in \ac{ioc} differs from the \ac{irl} literature mentioned above as it is explicitly designed for smooth problems.

One common framework for addressing \ac{ioc} problems with nonlinear dynamics and nonquadratic cost structures is bilevel optimization~\citep{mombaur2010ar,albrecht2011humanoids}.
Here, the \emph{outer} problem is a least squares or \ac{mle} problem in which demonstrations are matched with a nominal trajectory estimate and decision variables parameterize the objective of the underlying optimal control problem.
The \emph{inner} problem determines the nominal trajectory estimate as the optimizer of the ``forward'' (i.e., standard) optimal control problem for the outer problem's decision variables.
A key benefit of bilevel \ac{ioc} formulations is that they naturally adapt to settings with noise-corrupted partial state observations~\citep{albrecht2011humanoids}. %

Early bilevel formulations for \ac{ioc} utilize derivative-free optimization schemes to estimate the unknown objective parameters in order to avoid explicit differentiation of the solution to the inner optimal control problem~\citep{mombaur2010ar}.
That is, the inner solver is treated as a black-box mapping from cost parameters to optimal trajectories which is utilized by the outer solver to identify the unknown parameters using a suitable derivative-free method.
While black-box approaches can be simple to implement due to their modularity and lack of reliance on derivative information, they often suffer from a high sampling complexity~\citep{nocedal2006optimizationbook}. 
Since each sample in the context of black-box \ac{ioc} methods amounts to solving a full optimal control problem, such approaches remain intractable for scenarios with large state spaces or additional unknown parameters, such as unknown initial conditions.

Other works instead embed the \ac{kkt} conditions of the inner problem as constraints on the outer problem.
Since these techniques enforce only first-order necessary conditions of optimality, globally optimal observations are unnecessary and locally optimal demonstrations suffice.
Yet, a key computational difficulty of \ac{kkt}-constrained \ac{ioc} formulations is that they yield a nonconvex optimization problem due to decision variables in the outer problem appearing nonlinearly with inner problem variables in \ac{kkt} constraints. 
This occurs even in the relatively benign case of linear-quadratic \ac{ioc}.

In contrast to bilevel optimization formulations where necessary conditions of optimality are embedded as constraints, recent methods \citep{levine2012icml,englert2018ijrr,awasthi2019master,menner2020arxiv,jin2021inverse} minimize the residual of these conditions directly at the demonstrations.
Since the observed demonstration is assumed to satisfy any constraints of the underlying forward optimal control problem, this method can be formulated as fully unconstrained optimization.
Additionally, these residual formulations yield a \emph{convex} optimization problem if the class of objective functions is convex in the unknown parameters at the demonstration \citep{keshavarz2011isic, englert2018ijrr}. This condition holds in the common setting of linearly-parameterized objective functions.
\citet{levine2012icml} propose a variant of this approach that utilizes quadratic approximations of the reward model around demonstrations to derive optimality residuals in a maximum entropy framework.
\citet{englert2018ijrr} present an extensions of this method do accommodate inequality constraints on states and inputs. 
Much like \ac{kkt}-constrained formulations, these residual methods operate on locally optimal demonstrations.
However, an important limitation of residual methods is that they require observations of full state and input sequences.
More recently, \citet{menner2020arxiv} compared \ac{ioc} techniques based on \ac{kkt} constraints and residuals and demonstrated inferior performance of the latter even in problems with linear dynamics and quadratic target objectives.

Our work takes inspiration from the \ac{kkt}-constraint formulation for single-player \ac{ioc} as discussed by \citet{albrecht2011humanoids} and \citet{menner2020arxiv}.
While these works apply only to single-player settings, we utilize the necessary conditions for \acp{olne} \citep{basar1998gametheorybook} to generalize this approach to noncooperative multi-player scenarios.

\subsection{Multi-Player Inverse Dynamic Games}\label{sec:related-multi-player}

Many of the \ac{ioc} techniques discussed above have close analogues in the context of multi-player inverse dynamic games.

As in single-player \ac{ioc}, methods akin to black-box bilinear optimization have also been studied in the context of inverse games \citep{peters2020master,cleac2020arxiv}.
\citet{peters2020master} uses a particle-filtering technique for online estimation of human behavior parameters.
This work demonstrates the importance of inferring human behavior parameters for accurate prediction in interactive scenarios.
However, there, inference is limited to a single parameter and the work highlight the challenges associated with scaling this sampling based approach to high-dimensional latent parameter spaces.
\citet{cleac2020arxiv} employ a similar derivative-free filtering technique based on an \acl{ukf}.
While this approach drastically reduces the overall sample complexity, it still relies on exact observations of the state to reduce the required number of solutions to full dynamic games at the inner level.

Another line of research has put forth solution techniques for inverse games that follow from the residual methods outlined in \cref{sec:related-single-player} \citep{kopf2017ifac,rothfuss2017ifac,awasthi2020acc,inga2019arxiv}.
\citet{kopf2017ifac} study a special case of an inverse linear-quadratic game in which the equilibrium feedback strategies of all but one player are known.
This assumption reduces the estimation problem to single-player \ac{ioc} to which the residual methods discussed above can be applied directly.
\citet{rothfuss2017ifac} present a more general approach that does not exploit such special structure but instead minimizes the residual of the first-order necessary conditions for a local \ac{olne}.
\citet{inga2019arxiv} present a variant of this \ac{olne} residual method in a maximum entropy framework, generalizing the single-player \ac{ioc} algorithm proposed by \citet{levine2012icml}.
Recently, \citet{awasthi2020acc} also extended the \ac{olne} residual method of \citet{rothfuss2017ifac} to inverse games with state and input constraints. This approach extends that of \citet{englert2018ijrr} to noncooperative multi-player scenarios.

All of these inverse game \ac{kkt} residual methods share many properties with their single-player counterparts.
In particular, since they rely upon only local equilibrium criteria, they are able to recover player objectives even from local---rather than only global---equilibrium demonstrations. %
However, as in the single-player case, they rely upon observation of both state and input to evaluate the residuals.

In contrast to \ac{kkt} residual methods \citep{rothfuss2017ifac,awasthi2020acc,inga2019arxiv}, we enforce these conditions as constraints on a jointly estimated trajectory, rather than minimizing the residual of these conditions directly at the observation.
By maintaining a trajectory estimate in this manner, our method explicitly accounts for observation noise, partial state observability, and unobserved control inputs.
Furthermore, in contrast to black-box approaches to the inverse dynamic game problem~\citep{peters2020master,cleac2020arxiv}, our method does not require repeated solutions of the underlying forward game.
Moreover, our method returns a full forward game solution in addition to the estimated objective parameters for all players.

\section{Background: Open-Loop Nash Games}
\label{sec:background}

While this work is concerned with the \emph{inverse} game problem of learning objectives from observed behavior, we first provide a technical introduction to the theory of \emph{forward} open-loop dynamic Nash games.
These forward games correspond to the model that we seek to recover in this work.
Furthermore, as we shall discuss in \cref{sec:inverse_problem}, they may be used at the inner level of a bilevel optimization problem to formulate the inverse game problem.

As discussed in \cref{sec:intro}, dynamic games provide an expressive mathematical formalism for modeling the strategic interactions of multiple agents with differing objectives. Interested readers are directed to~\citep{basar1998gametheorybook} for a
more complete discussion.
We note that dynamic games afford a wide variety of
equilibrium concepts; our choice of open-loop Nash Equilibria in this work \revised{
captures scenarios in which players do not account for future information gains and instead commit to a sequence of control decisions \emph{a priori}.
These conditions may occur when occlusions \emph{prevent} future information gains or when bounded rationality causes players to \emph{ignore} them.
\Acp{olne} have been demonstrated to capture dynamic interaction when embedded in receding-horizon re-planning schemes~\citep{wang2019dars,cleac2020rss}.
Beyond that, restricting our attention to \acp{olne} engenders} computational
advantages \revised{which are} discussed below. \revised{Other choices of solution concept} are possible
and should be explored in future work.
Recent methods such as those of~\citet{di2019cdc} and~\citet{cleac2020rss} facilitate efficient solutions to the ``forward'' open-loop games \emph{given players' objectives a priori}.

\subsection{Preliminaries}
\label{sec:preliminaries}

Consider a game played between~$\numplayers$ players over discrete time-steps~$t
\in [\horizon] := \{1,\dots,\horizon\}$.
The game is comprised of three key components: dynamics, objectives (which are later presumed to be unknown in this work), and information structure.

We presume that the game is Markov with respect to state~$\state \in
\mbb{R}^\xdim$. That is, given each player's control input~$\control{i}~\in~\mbb{R}^{\udim{i}},~i~\in~[N]$, the state evolves according to the difference equation
\begin{equation}
  \label{eqn:dynamics}
  \state_{t+1}
  = \dynamics{t}(\state_t, \control{1}_t,\dots,\control{N}_t).
\end{equation}
For clarity, we shall introduce the following shorthand notation:
\begin{align*}
  \bstate &= (\state_1, \dots, \state_\horizon),\\
  \bcontrol{i} &= (\control{i}_1, \dots, \control{i}_\horizon),\\
  \bcontrol{}_t &= (\control{1}_t, \dots, \control{\numplayers}_t),\\
  \bcontrol{} &= (\bcontrol{1}, \dots, \bcontrol{\numplayers}).
\end{align*}
Observe that the state~$\state$ pertains to the entire game, not only to a
single player. In the examples presented in this paper,~$\state$ is simply the
concatenation of individual players' states, and correspondingly the dynamics
are independent for all players. However, this is not always the case and the methods developed here apply in the more general settings as well.

The objective of player $i$ is encoded by their distinct cost function~$\cost{i}$, which they seek to minimize.
This cost can in general depend upon the sequence of states and inputs for all players.\footnote{State and input constraints are also possible, although they complicate the notion of equilibrium solution.
Solution methods such as those of \citet{dirkse1995path} and \citet{laine2021arxiv} address constrained forward games.
The present paper readily extends to the constrained case; however, we neglect them for clarity of presentation.}
In this paper, we presume that objectives are expressed in time-additive form, as is common across the optimal control and reinforcement learning literature:
\begin{equation}
  \label{eqn:costs}
  \cost{i}(\bstate, \bcontrol{}) := \sum_{t=1}^\horizon
  \runningcost{i}_t(\state_t, \control{1}_t, \dots, \control{\numplayers}_t).
\end{equation}
Since the state trajectory~$\bstate$ follows \cref{eqn:dynamics}, these cost functions can also be written in terms of the inital condition~$\state_1$ and the sequence of control inputs for all players~$\bcontrol{}$.
For this reason, we shall also use the notation~$\cost{i}(\bcontrol{}; \state_1)$, and refer to the tuple of initial state, dynamics, and objectives as
\begin{equation}
  \label{eqn:game_tuple}
  \game := \left(\state_1, \{\dynamics{t}\}_{t \in [\horizon]}, \{\cost{i}\}_{i \in
      [\numplayers]}\right).
\end{equation}

Finally, the information structure of a dynamic game refers to the information
available to each player when they are required to make a decision at each time.
At time~$t$, then, \player{i}'s input is a function~$\strategy^i_t :
\infoset^i_t \to \mbb{R}^{\udim{i}}$, where~$\infoset^i_t$ is the set of
information available to \player{i} at time~$t$. In this paper, we consider
\emph{open-loop} information structures,~i.e., where~$\infoset^i_t =
\{\state_1\}$.\footnote{Recent work in solving forward games also considers
  \emph{feedback} information in which~$\infoset^i_t = \{\state_t\}$; see
  \citet{fridovich2020icra} and \citet{laine2021arxiv}.} In open-loop information, then, it
suffices for \player{i} to specify their input sequence~$\bcontrol{i}$ given
a fixed initial condition~$x_1$.
For this reason, we neglect a more detailed treatment of strategy spaces and information structure, and simply refer to the finite-dimensional sequence of control inputs for each player.

This characterization of a dynamic game is intentionally general.
Our solution methods will rely upon established numerical methods for smooth optimization, however, and as such we require the following assumption.
\begin{assumption} (Smoothness)
  \label{assume:smoothness}
  Dynamics~$\dynamics{}$ and objectives~$\cost{i}$ have well-defined second
  derivatives in all state and control variables, at all times and for all
  players.
\end{assumption}
Most physical systems of interest and
interactions thereof are naturally modeled in this way.
However, we note that, for example, hybrid dynamics such as those induced by contact do not satisfy this assumption.

We shall illustrate key concepts using a consistent ``running example''
throughout the paper.

\example{ Consider an~$\numplayers~=~2$-player \ac{lq} game---i.e., one in which
  dynamics~$\dynamics{t}$ are linear in state~$\state_t$ and control inputs~$\bcontrol{}_t$, and costs~$\cost{i}$ are quadratic in states and
  controls. Let each player independently follow the
  dynamics of a double integrator in the Cartesian plane. State~$\state =
  (p^1_x, p^1_y, \dot p^1_x, \dot p^1_y, p^2_x, p^2_y, \dot p^2_x, \dot p^2_y)$
  then evolves with inputs~$\control{i} = (\ddot p^i_x, \ddot p^i_y)$ according
  to
  \begin{align}
    \label{eqn:double_integrator}
    \state_{t+1} &= \overbrace{\begin{bmatrix}
      \tilde A & 0\\
      0 & \tilde A
    \end{bmatrix}}^{A} \state_t + \overbrace{\begin{bmatrix}
      \tilde B \\
      0
    \end{bmatrix}}^{B^1} \control{1}_t + \overbrace{\begin{bmatrix}
      0 \\
      \tilde B
    \end{bmatrix}}^{B^2} \control{2}_t,\\
                 &\textrm{where}~\tilde A = \begin{bmatrix}
                   1 & 0 & \Delta t & 0\\
                   0 & 1 & 0 & \Delta t\\
                   0 & 0 & 1 & 0\\
                   0 & 0 & 0 & 1
                 \end{bmatrix}, \tilde B = \begin{bmatrix}
                   0 & 0\\
                   0 & 0\\
                   \Delta t & 0\\
                   0 & \Delta t
                 \end{bmatrix}, \nonumber
  \end{align}
  and~$\Delta t$ is a uniform time discretization, e.g., 0.1s.
  Each player has a quadratic objective of the form
  \begin{equation}
  \label{eqn:quadratic_objectives}
    \cost{i} = \frac{1}{2} \sum_{t = 1}^\horizon \left( \costparams{i}_Q \state_t Q^i_t \state_t + \sum_{j = 1}^\numplayers \costparams{ij}_R \control{j \top}_t R^{ij}_t \control{j}_t \right).
  \end{equation}
  In this simple example,~$Q^i_t$ and~$R^{ij}_t$ are known, positive definite matrices encoding the preferences of each player.
  The scalars~$\costparams{i}_Q \in \mbb{R}$ and~$\costparams{ij}_R \in \mbb{R}$ \emph{weight} these known matrices.
  In this paper, we develop a technique to learn \emph{a priori} unknown parameters such as the costs weights above from both offline and online data.
  Note that this simple LQ game shall only serve to explain the general concepts of our method.
  For our experiments presented in \cref{sec:experiments}, we consider more complex problems with nonlinear dynamics and nonquadratic costs, such as the 5-player highway navigation problem shown in \cref{fig:frontfig}.
}

\subsection{The Nash Solution Concept}
\label{sec:nash}

Combining these components, each player $i$ in an open-loop dynamic game seeks
to solve the following optimization problem
\begin{subequations}
  \label{eqn:nash}
  \begin{numcases}{\forall i \in [\numplayers]}
    \label{eqn:coupled_optimization_problems}
    \min_{\bstate, \bcontrol{i}}~\cost{i}(\bcontrol{}; \state_1)\\
    \label{eqn:dynamics_constraint_t}
    \textrm{s.t.}~\state_{t+1} = \dynamics{t}(\state_t, \bcontrol{}_t), \forall
    t \in [T-1].
  \end{numcases}
\end{subequations}

There exist a variety of distinct solution concepts for such smooth open-loop
dynamic games. In this paper, we consider the well-known Nash equilibrium
concept, wherein no player has a unilateral incentive to change its strategy.
Mathematically, the Nash concept is defined as follows.
\begin{definition} (\Acl{olne})
  \label{def:nash}
  The strategies~$\bcontrol{*} := (\bcontrol{1*}, \dots,
  \bcontrol{\numplayers*})$ constitute an \acf{olne} in the game~$\game =
  \left(\state_1, \{\dynamics{t}\}_{t \in [T]}, \{\cost{i}\}_{i \in [\numplayers]}\right)$ if
  the following inequalities hold:
  \begin{equation}
    \label{eqn:nash_inequalities}
    \cost{i*} = \cost{i}(\bcontrol{*}; \state_1) \le \cost{i}\big((\bcontrol{i}, \bcontrol{-i*}); \state_1\big), \forall i \in [\numplayers].
  \end{equation}
  Here, we use the shorthand~$(\bcontrol{i}, \bcontrol{-i*})$ to indicate the
  collection of strategies in which \emph{only} \player{i}
  deviates from the Nash profile,~i.e.,~$\bcontrol{i} \ne \bcontrol{i*}$.
\end{definition}
Note that, at a Nash equilibrium, each player must \emph{independently} have no
incentive to deviate from its strategy.
Since players' objectives may generally conflict, the Nash concept encodes noncooperative, rational, and potentially selfish behavior.

Unfortunately, Nash equilibria are known to be very difficult to find in general \citep{daskalakis2009complexity}.
In this work, we seek only \emph{local} equilibria which satisfy the Nash conditions \cref{eqn:nash_inequalities} to first order.
That is, following similar approaches in both single-player \ac{ioc}
\citep{albrecht2011humanoids,englert2018ijrr} and forward/inverse open-loop games \citep{cleac2020rss,awasthi2019master}, we encode forward optimality 
via the corresponding first-order necessary conditions.
These
first-order necessary conditions are given by the union of the individual
players' \ac{kkt} conditions,~\ie,
\begin{multline}
  \label{eqn:forward_kkt_conditions}
  \mathbf{0} = \kktresidual(\bstate, \bcontrol{}, \bcostate{}) :=\\ \begin{bmatrix}
    \begin{rcases}
      \nabla_\bstate \cost{i} + \nabla_\bstate \capdynamics(\bstate, \bcontrol{})^\top \bcostate{i}\\
      \nabla_{\bcontrol{i}} \cost{i} +  \nabla_{\bcontrol{i}}
      \capdynamics(\bstate, \bcontrol{})^\top \bcostate{i}
    \end{rcases}
    \text{$\forall i \in [\numplayers]$}
    \\
    \capdynamics(\bstate, \bcontrol{})
  \end{bmatrix}.
\end{multline}
\noindent

Here, the first two block-rows are repeated for all players, and the function~$\capdynamics(\bstate, \bcontrol{})$ accumulates the dynamic constraints of \cref{eqn:dynamics_constraint_t} at all time steps, with the~$t^{\textnormal{th}}$ row given by
$\state_{t+1} - \dynamics{t}(\state_t, \control{1}_t, \dots,
\control{\numplayers}_t)$.
Note that we have also introduced costate variables~$\bcostate{i} :=
(\costate{i}_1, \dots, \costate{i}_{\horizon-1})$ for each player, with
$\costate{i}_t \in \mbb{R}^\xdim$ the Lagrange multiplier corresponding to \player{i}'s dynamics constraint in \cref{eqn:dynamics_constraint_t} at time step~$t$. Note that, as with control inputs, we use the notation~${\bcostate{} := (\bcostate{1}, \dots, \bcostate{\numplayers})}$.

\example{
Consider the two-player \ac{lq} example above with double integrator dynamics given by \cref{eqn:double_integrator} and quadratic objectives given by \cref{eqn:quadratic_objectives}.
The~$t^{\text{th}}$ block of the first row of \cref{eqn:forward_kkt_conditions} is given by
\begin{equation}
    \label{eqn:double_integrator_kkt_x}
    \mbf{0} = \costparams{i}_Q Q^i_t \state_t + \costate{i}_{t-1} - A^\top \costate{i}_t
\end{equation}
for \player{i}. Likewise, the~$t^{\text{th}}$ block of the second row of \cref{eqn:forward_kkt_conditions} for \player{i} is given by
\begin{equation}
    \label{eqn:double_integrator_kkt_u}
    \mbf{0} = \costparams{ii}_R R^{ii}_t \control{i}_t - B^{i\top} \costate{i}_t.
\end{equation}
Finally, the~$t^{\text{th}}$ block of the final row of \cref{eqn:forward_kkt_conditions} is given by
\begin{equation}
    \label{eqn:double_integrator_kkt_dyn}
    \mbf{0} = \state_{t+1} - A \state_t - B^1 \control{1}_t - B^2 \control{2}_t.
\end{equation}
}

Computationally, the \ac{kkt} conditions of the forward game, given in \cref{eqn:forward_kkt_conditions}, are a set of, generally nonlinear, equality constraints in the variables~$\bstate, \bcontrol{}$, and~$\bcostate{}$.
To find a solution---that is, a root of~$\kktresidual(\bstate, \bcontrol{}, \bcostate{})$---we may employ a root-finding algorithm such as a variant of Newton's method \cite[Chapter 11]{nocedal2006optimizationbook}. This is the approach taken by, e.g., \citet{cleac2020rss}.

\example{
For our \ac{lq} example, it can be seen that a single step of Newton's method on~$\kktresidual(\cdot)$ amounts to the well-known Riccati solution to an \emph{open-loop} \ac{lq} game \cite[Chapter 6]{basar1998gametheorybook}.\footnote{Note that this Newton step \emph{differs} from that given by the Riccati solution to a \emph{feedback} \ac{lq} game.}
}

\section{Problem Setup}\label{sec:inverse_problem}

Solving a \emph{forward} Nash game amounts to identifying optimal strategies for all players, provided a priori knowledge of their objectives~$\cost{i}$.
By contrast, in this work we are concerned with the \emph{inverse} Nash problem,~i.e., that of identifying players' objectives which explain their observed behavior.
To develop the inverse Nash problem, here we shall presume that learning occurs offline, given a sequence of noisy, partial observations of all players' state.
The method we develop for this setting, however, is amenable to trajectory prediction and online, receding horizon operation as discussed in~\cref{sec:online_learning}.

We formulate the inverse Nash problem as one of offline learning, in which players' objectives belong to a known parametric function class.
To that end, we make the following assumption.

\begin{assumption} (Parametric objectives)
\label{assume:parametric}
\player{i}'s cost function is fully described by a vector of parameters~$\costparams{i} \in \mbb{R}^{\costparamdim{i}}$.
That is,~$\cost{i}(\cdot; \costparams{i}) \equiv \sum_{t=1}^\horizon \runningcost{i}_t(\state_t, \control{1}_t, \dots, \control{\numplayers}_t; \costparams{i})$.
\end{assumption}

Recalling \cref{assume:smoothness}, the functions~$\runningcost{i}_t(\cdot; \costparams{i})$ have well-defined derivatives in states~$\state_t$ and controls~$\control{i}$.
We shall also extend this smoothness assumption to include the parameters themselves.

\begin{assumption} (Smoothness in parameter space)
\label{assume:smooth_params}
Extending \cref{assume:smoothness}, we require that stage cost functions~$\runningcost{i}_t(\cdot; \costparams{i})$ have well-defined first- and second-derivatives with respect to the parameter vector~$\costparams{i}$.
\end{assumption}

This smoothness assumption is quite general.
For example, players' stage costs~$\runningcost{i}_t(\cdot; \costparams{i})$ may be encoded as arbitrary function approximators such as artificial neural networks.
In this work, we choose a more interpretable (though less flexible) parametric structure; \revised{we defer an investigation of more general cost structures for future work}.
In particular, the examples considered here use a \emph{linearly-parameterized} structure in which~$\runningcost{i}_t(\cdot; \costparams{i})$ is a linear function of~$\costparams{i}$,~i.e.,~$\runningcost{i}_t(\cdot; \costparams{i}) \equiv \costparams{i\top} \tilde{\runningcost{}}^i_t(\cdot)$ for some set of potentially nonlinear basis functions~$\tilde{\runningcost{}}^i_t(\cdot)$.
By incorporating appropriate domain-specific knowledge, however, these relatively simple cost structures are able to encapsulate complex, strategic interactions such as the highway lane changes of \cref{fig:frontfig}.

\example{
Recall the quadratic objectives of \cref{eqn:quadratic_objectives}, and take cost parameters~$\costparams{i} = (\costparams{i}_Q, \costparams{ij}_R)_{j \in [\numplayers]}$.
Observe, therefore, that \player{i}'s objective depends linearly upon its cost parameters~$\costparams{i}$.
}

Thus equipped, the objective learning problem reduces to maximizing the likelihood of a sequence of partial state observations~$\bobservation := (\observation_1, \ldots, \observation_T)$ for the parametric class of games~$\game(\costparams{}) = \left(\state_1, \dynamics{}, \{\cost{(i)}(\,\cdot\,; \costparams{(i)})\}_{i \in [\numplayers]}\right)$.
Formally, we seek to solve a problem of the form
\begin{subequations}
\label{eqn:inverse_problem}
\begin{align}
\label{eqn:inverse_objective}
    \max_{\costparams{}, \bstate, \bcontrol{}} \quad& p(\bobservation \given \bstate, \bcontrol{})\\
\label{eqn:inverse_olne_constraint}
    \textrm{s.t.} \quad& (\bstate, \bcontrol{}) \text{ is an OLNE of } \game(\costparams{})\\
\label{eqn:inverse_dynamics_constraints}
                       & (\bstate, \bcontrol{}) \text{ is dynamically feasible under } f,
\end{align}
\end{subequations}
where~$\costparams{}$ aggregates all players' cost parameters,~i.e.,~$\costparams{} := (\costparams{1}, \ldots, \costparams{\numplayers})$, and~$p(\bobservation\given \bstate, \bcontrol{})$ constitutes a known observation likelihood, or measurement, model. %

\begin{remark} (Initial state)
Observe that~$\state_1$ is an explicit decision variable in \cref{eqn:inverse_problem}, whereas it represents a constant (known) initial condition in the forward game problem discussed in \cref{sec:background}.
This reflects the fact that the state trajectory, including initial conditions, must be estimated as part of the inverse problem.
As we shall see, estimating the state trajectory jointly with the cost parameters allows our method to be less sensitive to observation noise.
\end{remark}

This measurement model is arbitrary, though, following \cref{assume:smoothness} and \cref{assume:smooth_params}, it must be smooth.
In the simplest instance, we may receive an exact measurement of the sequence of states and inputs for all players.
In that case, the measurement model~$p(\bobservation\given \bstate, \bcontrol{})$ reduces to a Dirac delta function.
More generally,~$p(\bobservation\given \bstate, \bcontrol{})$ may be an arbitrary smooth probability density function, making our formulation amenable to realistic sensors such as cameras or LiDARs.

Prior work in both single-player \ac{ioc}, such as that of \citet{englert2018ijrr}, and inverse games, such as those of \citet{awasthi2020acc} and \citet{rothfuss2017ifac}, presumes a degenerate measurement model in which states and controls are observed directly without any noise.
When perfect observations are unavailable, these methods naturally extend by first estimating a sequence of likely states and controls (a standard nonlinear filtering problem).
In \cref{sec:baseline}, we describe these sequential estimation methods in greater detail.
In contrast, our formulation given in \cref{eqn:inverse_problem} encodes a coupled estimation problem in which states, control inputs, and cost parameters must all be estimated simultaneously.
Thus, our method exploits the additional coupling imposed by the Nash equilibrium constraints onto the unknowns.
In \cref{sec:experiments}, we conduct a series of Monte Carlo experiments to quantify the advantages afforded by simultaneous learning over sequential estimation.

\section{Equilibrium-Constrained Cost Learning} %
\label{sec:solution}

Here we present our core contribution, a mathematical formulation of objective inference in multi-agent, noncooperative games.
We express this problem as a nonconvex optimization problem with equilibrium constraints, which we relax into a standard-format equality-constrained nonlinear program.

\subsection{Offline Learning}
\label{sec:offline_learning}

We first consider the problem of learning each player's objective from previously recorded data of prior interactions, \emph{offline}.

\cref{eqn:inverse_problem} is a mathematical program with equilibrium constraints \citep{luo1996mathematical, ferris2005mathematical}, with the nested equilibrium conditions of \cref{eqn:inverse_olne_constraint} encoding the Nash inequalities of \cref{def:nash}.
Equilibrium constraints generalize bilevel programming, and computational approaches tend to be less mature than those for standard-form (in)equality-constrained programming.

We relax the equilibrium constraint of \cref{eqn:inverse_olne_constraint} by replacing it with its \ac{kkt} conditions,~i.e., by substituting \cref{eqn:forward_kkt_conditions}.
This yields:
\begin{subequations}
\label{eqn:inverse_approach}
\begin{align}
\label{eqn:inverse_approach_objective}
    \max_{\costparams{}, \bstate, \bcontrol{}, \bcostate{}} \quad& p(\bobservation \given \bstate, \bcontrol{})\\
\label{eqn:inverse_kkt_constraints}
    \textrm{s.t.} \quad& \kktresidual(\bstate, \bcontrol{}, \bcostate{}; \costparams{})  = \mathbf{0}.
\end{align}
\end{subequations}
Here, we have explicitly written the \ac{kkt} conditions from \cref{eqn:forward_kkt_conditions} in terms of the cost parameters~$\costparams{}$.
Additionally, observe that in \cref{eqn:inverse_approach}, the costates~$\bcostate{}$ required to evaluate the \ac{kkt} conditions~$\kktresidual(\cdot; \costparams{})$ appear as \emph{additional primal variables}.
The constraints of \cref{eqn:inverse_kkt_constraints} will be assigned their own Lagrange multipliers, which are distinct from the original costates.
By letting states, control inputs, and costates be primal variables, the \ac{kkt} conditions~$\kktresidual(\cdot)$ do not depend explicitly upon the observations~$\bobservation$.
Thus, solving \cref{eqn:inverse_approach} does not require complete state or input observations; rather, the equilibrium constraints of \cref{eqn:inverse_kkt_constraints} allow us to reconstruct this missing information while we estimate cost parameters~$\costparams{}$, simultaneously.
Several remarks are in order.

\begin{remark} (Multiple observed trajectories)
We have developed \cref{eqn:inverse_approach} for the setting in which a single trajectory~$(\bstate, \bcontrol{})$ has been observed, yielding a measurement sequence~$\bobservation$.
However, our approach affords straightforward extension to settings in which player's objectives are learned from multiple demonstrations.
In this instance, the primal variables~$(\bstate, \bcontrol{}, \bcostate{})$ would be replicated for all trajectories, although the cost parameters~$\costparams{}$ would be shared.
The objective given by \cref{eqn:inverse_approach_objective} would be replaced by the joint probability of all measurements  conditioned on all underlying trajectories, and the equilibrium constraints in \cref{eqn:inverse_kkt_constraints} would be concatenated for all trajectories.
\end{remark}

\begin{remark} (Regularizing parameters)
\label{remark:regularization}
Depending upon the parametric structure of players' objectives~$\cost{i}(\cdot; \costparams{i})$, and hence the structure of \ac{kkt} residual~$\kktresidual(\cdot; \costparams{})$, it can be critical to regularize and/or constrain cost parameters.
For example, if there exists a choice of~$\costparams{i}$ for $\player{i}$ such that~$\cost{i}(\bstate, \bcontrol{}; \costparams{i})$ is constant for all dynamically-feasible trajectories~$(\bstate, \bcontrol{})$, then every such trajectory would satisfy the equilibrium constraint of \cref{eqn:inverse_olne_constraint}.
Such choices of~$\costparams{}$ must be avoided, \eg, by regularizing or otherwise constraining parameters.
\end{remark}

\example{
Following \cref{remark:regularization}, we constrain the parameters~$\costparams{i} \ge \costparamfloor > 0$.
Moreover, to account for scale invariance, we constrain their sum to unity,~i.e.,~$\sum_{i\in[\numplayers]} \left(\costparams{i}_Q + \sum_{j\in[\numplayers]} \costparams{ij}_R\right) = 1$.
}

\subsubsection{Least Squares}
\label{sec:least_squares}

A common observation model \mbox{$p(\bobservation \given \bstate, \bcontrol{})$} is the \ac{awgn} model.
Here, each observation~$\observation_t$ depends only upon the current state~$\state_t$ and control inputs~$\bcontrol{}_t$,~i.e.,
\begin{align}\label{eqn:awgn_observation}
    \observation_t = \obsmap_t(\state_t, \bcontrol{}_t) + \noise_t, %
\end{align}
\noindent where the (potentially nonlinear) function~$\obsmap_t$ computes the expected measurement, and~$\noise_t$ is a zero-mean Gaussian white noise process with known covariance, i.e.,~$\noise_t\sim\mathcal{N}(0, \noisecov_t)$.
In this case, following standard methods in maximum likelihood estimation \citep{gallager2013stochastic}, it is straightforward to express the maximization in \cref{eqn:inverse_approach_objective} as nonlinear least squares by taking the negative logarithm of~${p(\bobservation \given \bstate, \bcontrol{})}$:
\begin{subequations}
\label{eqn:inverse_approach_lsq}
\begin{align}
\label{eqn:inverse_approach_objective_lsq}
    \min_{\costparams{}, \bstate, \bcontrol{}, \bcostate{}} \quad& \sum_{t=1}^{\horizon} \big(\observation_t - \obsmap_t(\state_t)\big)^\top \noisecov_t^{-1} \big(\observation_t - \obsmap_t(\state_t)\big)\\
\label{eqn:inverse_kkt_constraints_lsq}
    \textnormal{s.t.} \quad& \kktresidual(\bstate, \bcontrol{}, \bcostate{}; \costparams{})  = \mathbf{0}.
\end{align}
\end{subequations}

In summary, this inverse problem entails the following task:
Find those parameters $\theta$ for which the corresponding game solution generates expected observations near the observed data.
This formulation of the inverse game problem can be encoded using well-established numerical modeling languages such as CasADi \citep{andersson2019mpc} or JuMP \citep{dunning2017sirev}, and solved using off-the-shelf optimization routines such as IPOPT \citep{wachter2006jmp} or SNOPT \citep{gill2005sirev}.

\subsubsection{Problem Complexity}
\label{sec:complexity}

Let us examine the structure of the least squares problem in \cref{eqn:inverse_approach_lsq} more carefully.
In general, the observation map~$\obsmap_t(\cdot)$ and  \ac{kkt} conditions~$\kktresidual(\cdot; \cdot)$ may be arbitrarily nonlinear.
Therefore, without further structural assumptions, our formulation is an equality-constrained nonlinear least squares problem.
Due primarily to the nonlinearities in~$\kktresidual$, \cref{eqn:inverse_approach_lsq} is generally nonconvex.
Solution methods, therefore, may be sensitive to initial values of primal variables; we discuss a straightforward initialization scheme in \cref{sec:initialization}.

Perhaps surprisingly, this nonconvexity persists in the \ac{lq} setting of our running example, even when~$\obsmap_t(\cdot)$ is the identity.

\example{
Consider the \ac{lq} setting, with~$\costparams{i} = (\costparams{i}_Q, \costparams{ij}_R)_{j \in [\numplayers]}$ as before.
Let the observation map be the identity,~i.e.,~$\obsmap_t(\state_t) = \state_t$ and presume \ac{awgn}.
The resulting nonlinear least squares problem in \cref{eqn:inverse_approach_lsq} has constraints of the form given in \crefrange{eqn:double_integrator_kkt_x}{eqn:double_integrator_kkt_dyn}.
Let us consider the first of these constraints for a single time step~$t$ and \player{i}:
\begin{equation*}
    \mbf{0} = \costparams{i}_Q Q^i_t \state_t + \costate{i}_{t-1} - A^\top \costate{i}_t.
\end{equation*}
Recall that the decision variables in our formulation are~$(\costparams{}, \bstate, \bcontrol{}, \bcostate{})$.
Here, we see that~$\costparams{i}$ multiplies~$\state_t$.
At best, therefore, this constraint is a \emph{bilinear} equality, making the overall problem in \cref{eqn:inverse_approach_lsq} nonconvex even for this minimal inverse \ac{lq} game.
}

When we directly observe both state and control inputs without noise,~i.e., \mbox{$\observation_t \equiv (\state_t, \bcontrol{}_t)$}, these constraints become \emph{linear} even in the general non-\ac{lq} setting, so long as players' objectives are linearly parameterized.
In this setting, we may rewrite \cref{eqn:double_integrator_kkt_x} as
\begin{subequations}
\begin{align}
    \mbf{0} &= \nabla_{\state_t} \overbrace{\runningcost{i}_t(\state_t, \bcontrolt; \costparams{i})}^{\costparams{i\top} \tilde{\runningcost{}}^i_t(\cdot)} + \costate{i}_{t-1} - \nabla_{\state_t} \dynamics{t}(\state_t, \bcontrolt)^\top \costate{i}_t \\
    \label{eqn:linear_constraint_x}
    &= \costparams{i\top} \nabla_{\state_t} \tilde{\runningcost{}}^i_t(\state_t, \bcontrolt) + \costate{i}_{t-1} - \nabla_{\state_t} \dynamics{t}(\state_t, \bcontrolt)^\top \costate{i}_t.
\end{align}
\end{subequations}
With this observation model, then, the only decision variables are~$(\costparams{i}, \costate{i}_t, \costate{i}_{t-1})$, which all appear linearly.
Furthermore, the least squares objective in \cref{eqn:inverse_approach_objective_lsq} becomes unnecessary, since, by assumption, the measurements~$\bobservation$ already include the states~$\bstate$ exactly.
Incorporating these simplifications, the entire constrained least squares problem of \cref{eqn:inverse_approach_lsq} reduces to the problem
\begin{subequations}
\label{eqn:reduced_lsq}
\begin{align}
    \mathrm{find} \quad& \costparams{}, \bcostate{}\\
    \mathrm{s.t.} \quad& \mbf{0} = \costparams{i\top} \nabla_{\state_t} \tilde{\runningcost{}}^i_t(\state_t, \bcontrolt) + \costate{i}_{t-1} \nonumber\\
    &\qquad\qquad\qquad\quad\, -\nabla_{\state_t} \dynamics{t}(\state_t, \bcontrolt)^\top \costate{i}_t, \forall i, t\\
    & \mbf{0} = \costparams{i\top} \nabla_{\control{i}_t} \tilde{\runningcost{}}^i_t(\state_t, \bcontrolt) \nonumber\\
    &\qquad\qquad\qquad\quad\, - \nabla_{\control{i}_t} \dynamics{t}(\state_t, \bcontrolt)^\top \costate{i}_t, \forall i, t.
\end{align}
\end{subequations}

Because the constraints in \cref{eqn:reduced_lsq} are linear, the problem is equivalent to a linear system of equations.
Moreover, since the constraints are completely decoupled for each player, they may be solved separately and in parallel for all players to obtain cost parameters~$\costparams{i}$ and costates~$\bcostate{i}$.
This reduction forms the basis for the state-of-the-art in solving inverse dynamic games \citep{rothfuss2017ifac, awasthi2020acc}, which only apply in settings with perfect state and input observations.
To compare against these methods in more general settings that feature noise, unobserved inputs, and partial state measurements, we augment these methods with a sequential optimization procedure in \cref{sec:baseline}.
Comparative Monte Carlo studies of all approaches are presented in \cref{sec:experiments}.

\subsection{Online Learning}
\label{sec:online_learning}
While \cref{sec:offline_learning} estimates the objectives of interacting agents from recorded data \emph{offline}, our formulation for inverse Nash problems extends naturally to an \emph{online} learning setting; \ie, cost learning from observations of ongoing interactions.
As we shall discuss below, our method can perform online cost learning and trajectory prediction simultaneously, making it suitable for receding horizon applications.

\subsubsection{Learning with Prediction}
\label{sec:offline_learning_prediction}

Equipped with a tractable solution strategy for the setting of offline learning, we now consider a coupled \emph{prediction} and learning problem.
Similar problems have been considered in the single-agent setting by, e.g., \citep{jin2021inverse,mukadam2019steap}.
Here, we aim to learn the cost parameters~$\costparams{}$ from only a \emph{subset} of the game horizon;~i.e., we presume that observations~$\bobservation = (\observation_1, \dots, \observation_\obshorizon)$ where the observation horizon~$\obshorizon \leq \horizon$.
Despite this change, the original problem of \cref{eqn:inverse_problem} remains effectively unchanged; only the objective has changed.
In particular, by substituting the \ac{kkt} conditions for an \ac{olne} in place of the original equilibrium constraint as in \cref{eqn:inverse_approach}, and making \ac{awgn} assumptions, we recover a variant of the constrained least squares formulation of \cref{eqn:inverse_approach_lsq}:
\begin{subequations}
\label{eqn:inverse_approach_prediction_lsq}
\begin{align}
\label{eqn:inverse_approach_prediction_objective_lsq}
    \min_{\costparams{}, \bstate, \bcontrol{}, \bcostate{}} \quad& \sum_{t=1}^{{\color{red}{\obshorizon}}} \big(\observation_t - \obsmap_t(\state_t)\big)^\top \noisecov_t^{-1} \big(\observation_t - \obsmap_t(\state_t)\big)\\
\label{eqn:inverse_kkt_prediction_constraints_lsq}
    \textnormal{s.t.} \quad& \kktresidual(\bstate, \bcontrol{}, \bcostate{}; \costparams{})  = \mathbf{0}.
\end{align}
\end{subequations}
Note that the upper limit of addition is~$\obshorizon$, rather than~$\horizon$ as in \cref{eqn:inverse_approach_objective_lsq}, while the \ac{olne} \ac{kkt} conditions in \cref{eqn:inverse_kkt_prediction_constraints_lsq} depend upon states, inputs, and costates for all times~$t \in \{1, \dots, \obshorizon, \dots, \horizon\}$.

\begin{figure*}[t]
    \centering
    \subfigure[Ours\label{fig:ours-pipeline}]{\includegraphics[scale=0.15]{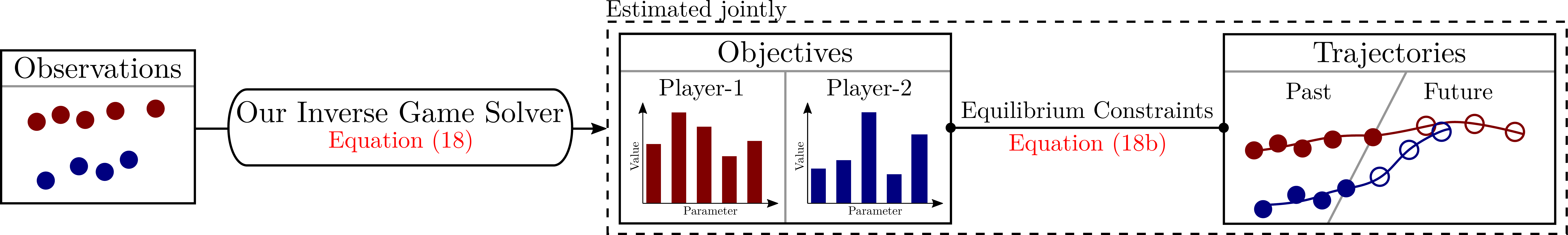}}
    \subfigure[Baseline\label{fig:baseline-pipeline}]{\includegraphics[scale=0.15]{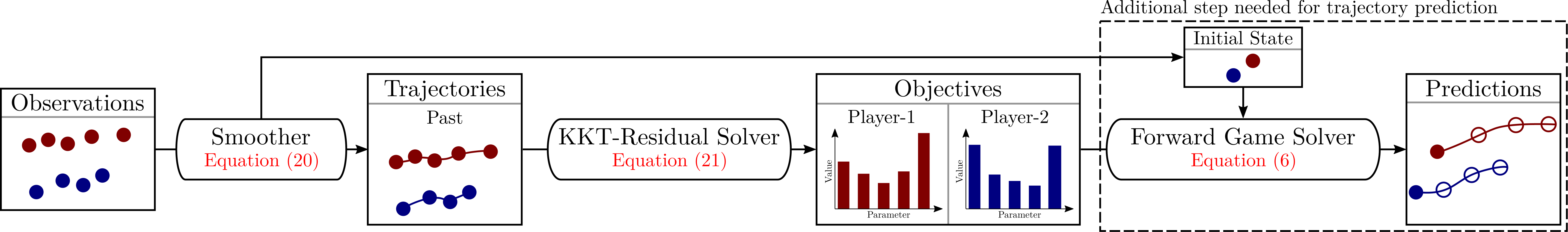}}
    \caption{Schematic overview of inverse game solvers set up for online operation. (a) Our method computes player's objectives, state estimates, and trajectory predictions jointly. (b) The baseline requires full knowledge of states and inputs and therefore must preprocess raw observations before it can estimate players' objectives. In order to generate trajectory predictions, the baseline must solve an additional forward game formulated over the estimated initial states and objectives.}
    \label{fig:baseline-pipeline}
\end{figure*}

Despite the similarities between this problem and \cref{eqn:inverse_approach_lsq}, the Nash trajectory~$(\bstate^*, \bcontrol{*})$, which emerges as a solution affords a new interpretation.
In particular, for times~$t \le \obshorizon$ these equilibrium states and controls constitute filtered estimates of the observed quantities~$\bobservation$, while for times~$t > \obshorizon$ they represent \emph{predictions} of the future.
Importantly, however, extending trajectories beyond the observation horizon~$\obshorizon$ adds additional constraints to \cref{eqn:inverse_approach_lsq}.
This ability to incorporate future, unobserved states makes the method more robust and data efficient when only a fraction of the game horizon is observed.
Consequently, this formulation can be employed for online learning in scenarios of ongoing interactions.
We provide a detailed empirical analysis of this setting in \cref{sec:detailed_offline_prediction_experiments}.
A summary of this variant of our inverse game solver is provided in \cref{fig:ours-pipeline}.

\subsubsection{Receding Horizon Learning}
\label{sec:receding_horizon}

Our method is directly amenable to receding horizon, online operation.
Here, we suppose that the agents interact over the half-open time-interval~$t \in \{1, \dots, \obshorizon, \dots, \infty\}$, and that observations exist for~$t \le \obshorizon$.
Here,~$\obshorizon$ may be interpreted as the current time and, as time elapses, both~$\obshorizon$ and the overall prediction horizon~$\horizon$ increase accordingly.
Unfortunately, however, increasing the overall problem horizon increases the number of variables in \cref{eqn:inverse_problem}, eventually making the problem intractable.

To simplify matters, we approximate the learning problem at each instant by neglecting all times outside the interval~$\{\obshorizon - s_\mathrm{o}, \dots, \obshorizon, \dots, \obshorizon + s_\mathrm{p} \}$, where~$s_\mathrm{o}$ is the length of a fixed-lag buffer of past observations, and $s_\mathrm{p}$ is the horizon of future state predictions.
In this setting, the total number of variables remains constant (since the length of this interval is constant), rendering \cref{eqn:inverse_problem} tractable to solve online.
More precisely, at time~$\obshorizon$ (and under \ac{awgn} assumptions), we solve a modified version of \cref{eqn:inverse_approach_prediction_lsq}
\begin{subequations}
\label{eqn:inverse_approach_online_lsq}
\begin{align}
\label{eqn:inverse_approach_online_objective_lsq}
    \min_{\costparams{}, \bstate, \bcontrol{}, \bcostate{}} \quad& \sum_{t={\color{red}{\obshorizon - s_\mathrm{o}}}}^{{\color{red}{\obshorizon}}} \big(\observation_t - \obsmap_t(\state_t)\big)^\top \noisecov_t^{-1} \big(\observation_t - \obsmap_t(\state_t)\big)\\
\label{eqn:inverse_kkt_online_constraints_lsq}
    \textnormal{s.t.} \quad& \kktresidual(\bstate, \bcontrol{}, \bcostate{}; \costparams{})  = \mathbf{0},
\end{align}
\end{subequations}
where the \ac{kkt} constraint~$\kktresidual(\cdot)$ is understood to depend upon times~$t \in \{\obshorizon - s_\mathrm{o}, \dots, \obshorizon, \dots, \obshorizon + s_\mathrm{p}\}$ and states, control inputs, and costates are also limited to that interval.
At each later time, we solve a problem with identical structure, with the understanding that~$\obshorizon$ will have changed to reflect the elapsed time.
In effect, this procedure amounts to simultaneous fixed-lag smoothing and receding-horizon prediction.
We simulate this online learning procedure in \cref{sec:online_learning_experiments}.

\section{Baseline}
\label{sec:baseline}

Recall the discussion of \cref{sec:complexity}, in which we show that---with noiseless observations of states~$\bstate$ and controls~$\bcontrol{}$, and linear cost parameterization~$\runningcost{i}_t(\cdot; \costparams{i}) \equiv \costparams{i\top} \tilde{\runningcost{}}^i_t(\cdot)$---our formulation reduces to the linear system of equations of \cref{eqn:reduced_lsq}.
This reduction underlies state of the art methods for learning the objectives of players in games \citep{rothfuss2017ifac, awasthi2020acc}.
Therefore, such methods unfortunately require noiseless observations of the full state and input sequences for all players.
In contrast, our approach in \cref{eqn:inverse_approach} is amenable to noisy, partial observations.

\subsection{Recovering Unobserved Variables}
\label{sec:initialization}

To provide a meaningful comparison between our proposed technique and the state-of-the-art in settings with imperfect observations, we augment \citep{rothfuss2017ifac,awasthi2020acc} with a pre-processing to estimate unobserved states and inputs.
To that end, we solve the following relaxed version of \cref{eqn:inverse_approach}:
\begin{subequations}
\label{eqn:pre_solve}
\begin{align}
\label{eqn:pre_solve_objective}
\tilde\bstate, \tilde{\bcontrol{}} := \arg\max_{\bstate, \bcontrol{}} \quad & p(\bobservation \given \bstate, \bcontrol{})\\
\label{eqn:pre_solve_constraint}
\textrm{s.t.}  \quad& \capdynamics(\bstate, \bcontrol{}) = \mbf{0}.
\end{align}
\end{subequations}
As in \cref{sec:least_squares}, under a \ac{awgn} assumption \cref{eqn:pre_solve} becomes equality-constrained nonlinear least squares.
However, unlike \cref{eqn:inverse_approach_lsq}, we have neglected the first two rows of the equilibrium constraint given in \cref{eqn:forward_kkt_conditions}.
That is, \cref{eqn:pre_solve} computes a maximum likelihood estimate of states and inputs irrespective of the underlying game structure.

The solution of this smoothing problem is used as an estimate of states an inputs when the baseline is employed in partially observed settings.
Beyond that, the same procedure serves as simple, yet effective initialization scheme for our method to tackle issues of non-convexity discussed in \cref{sec:complexity}.

\subsection{Minimizing KKT Residuals}
\label{sec:kkt_residuals}

Like our proposed method, the state-of-the-art methods developed by \citet{rothfuss2017ifac} and \citet{awasthi2020acc} use the forward game's \ac{kkt} conditions to measure the quality of a set of cost parameters~$\costparams{}$.
While we compare to this derivative-based, \ac{kkt} condition approach, we note that other approaches outlined in \cref{sec:related-multi-player} such as \citep{cleac2020arxiv} utilize black-box optimization methods and do not require or exploit derivative information.
These significant algorithmic differences---and the resulting differences in sample complexity, locality of solutions, etc.---make a direct comparison difficult to interpret.

Specifically, the \ac{kkt} residual method of \citep{awasthi2020acc, rothfuss2017ifac} fixes the state and input sequences to their observed---or in our case, estimated via \cref{eqn:pre_solve}---values.
Fixing these variables, however, the resulting linearly-constrained satisfiability problem of \cref{eqn:reduced_lsq} may be infeasible, depending upon the parametric structure of costs~$\runningcost{i}_t(\cdot; \costparams{i})$.
In lieu, state-of-the-art approaches minimize the \ac{kkt} residual itself, i.e.,
\begin{align}
\label{eqn:kkt_residual_baseline}
    \min_{\costparams{}, \bcostate{}} \norm{\kktresidual(\tilde\bstate, \tilde{\bcontrol{}}, \bcostate{}; \costparams{})}_2^2.
\end{align}
In prior work \citep{awasthi2020acc, rothfuss2017ifac},~$\tilde\bstate$ and~$\tilde{\bcontrol{}}$ are assumed to be directly observed.
As discussed in \cref{sec:initialization}, here we presume they are the results of the pre-processing step given in \cref{eqn:pre_solve}.
Additionally, like the linear system of equations in \cref{eqn:reduced_lsq}, the only decision variables here are the objective parameters~$\costparams{}$ and the costates~$\bcostate{}$.
In effect, the baseline does \emph{not} refine the state and input estimates given by the pre-processing step of \cref{eqn:pre_solve}.
Furthermore, as in \cref{eqn:reduced_lsq}, the problem may be decomposed into separate problems for each player and solved in parallel.
In essence, then, this \ac{kkt} residual formulation neglects the coupling between players' actions which is encoded in the equilibrium conditions; computationally, it reduces to solving separate \ac{ioc} problems for each player neglecting game-theoretic interactions with others.

A schematic overview of this baseline approach is depicted in \cref{fig:baseline-pipeline}.
By first estimating the states~$\bstate$ and inputs~$\bcontrol{}$ from measurements~$\bobservation$, and only afterward learning the cost parameters~$\costparams{}$ and associated costates~$\bcostate{}$, the \ac{kkt} residual method can be thought of as a sequential decomposition of our approach.
By contrast, our formulation maintains~$(\bstate, \bcontrol{})$ as decision variables and refines the initial guess of~$(\tilde{\bstate}, \tilde{\bcontrol{}})$ by  \emph{identifying all variables simultaneously}.

\section[Experiments]{Experiments\protect\footnote{Some result figures and descriptions are drawn from the earlier conference version of this work \citep{peters2021rss}.}}
\label{sec:experiments}

In this work, we develop a technique for learning players' objectives in continuous dynamic games from noise-corrupted, partial state observations.
We conduct a series of Monte Carlo studies to examine the relative performance of our proposed methods and the \ac{kkt} residual baseline in both offline and online learning settings.

\subsection{Experimental Setup}
\label{sec:experimental_setup}

We implement our proposed approach as well as the \ac{kkt} residual baseline of \cite{rothfuss2017ifac} in the Julia programming language \citep{bezanson2017sirev}, using the mathematical modeling framework JuMP \citep{dunning2017sirev}.
As a consequence, our implementation encodes an abstract description of \cref{eqn:inverse_approach}, making it straightforward to use in concert with a variety of optimization routines.
In this work, we use the open source COIN-OR IPOPT algorithm \citep{wachter2006jmp}.
The source code for our implementation is publicly available.\footnote{\href{https://github.com/PRBonn/PartiallyObservedInverseGames.jl}{https://github.com/PRBonn/PartiallyObservedInverseGames.jl}}

To evaluate the relative performance of our proposed approach with the \ac{kkt} residual baseline, we perform several Monte Carlo studies.
The details of these studies are described below.
However, all of these studies share the following overall setup: we fix a cost parameterization for each player, find corresponding \ac{olne} trajectories as roots of \cref{eqn:forward_kkt_conditions} using the well-known \ac{ibr} algorithm \cite{wang2019dars}, and simulate noisy observations thereof with \acf{awgn} as in \cref{eqn:awgn_observation}.
Each study then presents samples across a different problem parameter to test the sensitivity of both approaches to observation noise (\cref{sec:detailed_offline_experiments,sec:scaling_offline_experiments}) and unobserved time-steps (\cref{sec:detailed_offline_prediction_experiments}) in two different problem settings.

In each of the studies below, we consider~$\numplayers$ vehicles navigating traffic, and instantiate game dynamics and player objectives as follows.
Each vehicle has its own state~$\state^i$ such that the global game state is concatenated as~$\state =
(\state^1, \dots, \state^\numplayers)$. Further, each vehicle follows unicycle dynamics at time
discretization~$\Delta t$:
\begin{equation}
\label{eqn:runexp-dynamics}
\state_{t+1}^i =
\begin{cases}
  \textnormal{\emph{(x-position)}}~\position_{x,t+1}^i &= \position_{x,t}^i + \Delta t ~\speed_t^i \cos \heading_t^i\\
  \textnormal{\emph{(y-position)}}~\hfill\position_{y,t+1}^i &= \position_{y,t}^i + \Delta t ~\speed_t^i \sin \heading_t^i\\
  \textnormal{\emph{(heading)}}~\hfill\heading_{t+1}^i &= \heading_t^i + \Delta t ~\yawrate_t^i\\
  \textnormal{\emph{(speed)}}~\quad\hfill\speed_{t+1}^i &= \speed_t^i +
  \Delta t ~\acceleration_t^i,
\end{cases}
\end{equation}
where~$\control{i}_t = (\yawrate_t, \acceleration_t)$ includes the yaw rate and
longitudinal acceleration. Finally, each player's objective is characterized
by a stage cost~$\runningcost{i}_t$ which is a weighted sum of several
basis functions, i.e.,
\begin{subequations}
\label{eqn:runexp-cost}
\begin{numcases}{\runningcost{i}_t = \sum_{\ell = 1}^5 \weight^i_\ell
    \runningcost{i}_{\ell,t}}
  \runningcost{i}_{1,t} = \indicator(t \ge T - t_{\textnormal{goal}}) \goalmap(\state_t^i, \state_{\textnormal{goal}}^i)  \label{eqn:runexp-goal-cost}\\
  \runningcost{i}_{2,t} = -\sum_{j\ne i}\log(\|\position_i - \position_{j}\|_2^2)\label{eqn:runexp-prox-cost}\\
  g_{3,t}^i = (v^i)^2\label{eqn:runexp-vel-cost}\\
  \runningcost{i}_{4,t} = (\yawrate_t^i)^2 \label{eqn:runexp-yawrate-cost}\\
  \runningcost{i}_{5,t} =
  (\acceleration_t^i)^2 \label{eqn:runexp-accel-cost}.
\end{numcases}
\end{subequations}
Here, the cost parameters~$\costparams{i} = (\weight_\ell^i)_{\ell \in [5]}, \weight_\ell^i \in \mbb{R}_+$ are positive weights for each cost
component. Further,~$\position_i = (p^i_x, p^i_y)$ denotes the planar position of \player{i}, and~$\goalmap(\cdot, \cdot)$ is an arbitrary distance
mapping. For example, we may choose~$\goalmap(\state_t^i,
\state^i_{\textnormal{goal}}) = \|\position_{t}^i -
\position_{\textnormal{goal}}^i\|_2^2$ to compute squared distance from a fixed goal
position. 
Note, however, that this map is generic and can also be used to encode more complex goal-reaching specifications as in the highway lane-changing example depicted in \cref{fig:frontfig}. 
Taken together, the basis functions encode the following aspects of each player's
preferences:
\begin{enumerate}
\item Be close to the goal state within the last~$t_{\textnormal{goal}}$ time steps (\ref{eqn:runexp-goal-cost})
\item Avoid close proximity to other vehicles (\ref{eqn:runexp-prox-cost})
\item Avoid high speeds (\ref{eqn:runexp-vel-cost})
\item Avoid large control efforts (\ref{eqn:runexp-yawrate-cost}, \ref{eqn:runexp-accel-cost})
\end{enumerate}
Games of this form are inherently noncooperative since players must compete to reach
their own goals efficiently while avoiding collision with one another. Hence,
they must negotiate these conflicting objectives and thereby find an
equilibrium of the underlying game. 

In all of the Monte Carlo studies, we evaluate the approaches for two different noisy observation models~$\obsmap^\textrm{full}_{t}$ and~$\obsmap^\textrm{partial}_{t}$. 
In~$\obsmap^\textrm{full}_{t}(\state_t) := \state_t$, estimators observe the \emph{full} state, and in~$\obsmap^\textrm{partial}_{t}(\state_t) := (\position_t^1, \heading^1_t, \dots, \position_t^\numplayers, \heading^\numplayers_t)$, estimators observe the position and heading but not the speed of each agent; \ie, they receive a \emph{partial} state observation.

\subsection{Detailed Analysis of a 2-Player Game}
\label{sec:detailed_experiments}

We first study the performance of our method in a simplified,~$\numplayers = 2$-player game.
This set of experiments demonstrates the performance gap of our approach and the \ac{kkt} residual baseline in methods in a conceptually simple and easily interpretable scenario.
Here, the game dynamics are given as in \cref{eqn:runexp-dynamics}, and player objectives are parameterized as in \cref{eqn:runexp-cost}.
In particular, we let~${\goalmap(\state_t^i,
\state^i_{\textnormal{goal}}) = \|\position_{t}^i -
\position_{\textnormal{goal}}^i\|_2^2}$.
In summary, therefore, each vehicle wishes to reach a fixed, known goal position in the plane while avoiding collision with the other.
\subsubsection{Offline Learning}
\label{sec:detailed_offline_experiments}
We begin by studying both our method's and the baseline's ability to infer the unknown objective parameters~$\costparams{}$, as developed in \cref{sec:offline_learning}.
To do so, we conduct a Monte Carlo study for the aforementioned 2-player collision-avoidance application.

We generate 40 random observation sequences at each of 22 different levels of isotropic observation noise.
For each of the resulting 880 observation sequences we run both our method and the baseline to recover estimates of weights~\mbox{$\costparams{i} = (\weight_\ell^i)_{\ell \in [5]}$} for each player. %
Note that in this offline setting both methods learn these objective parameters from noisy observations of a single, complete game trajectory.
That is, each estimate relies upon \SI{25}{\second} of simulated interaction history from a single scenario.

\begin{figure}
    \centering
    \subfigure[Parameter estimation\label{fig:runexmp-estimator_parameter_error}]{
    \includegraphics[scale=\vegascale]{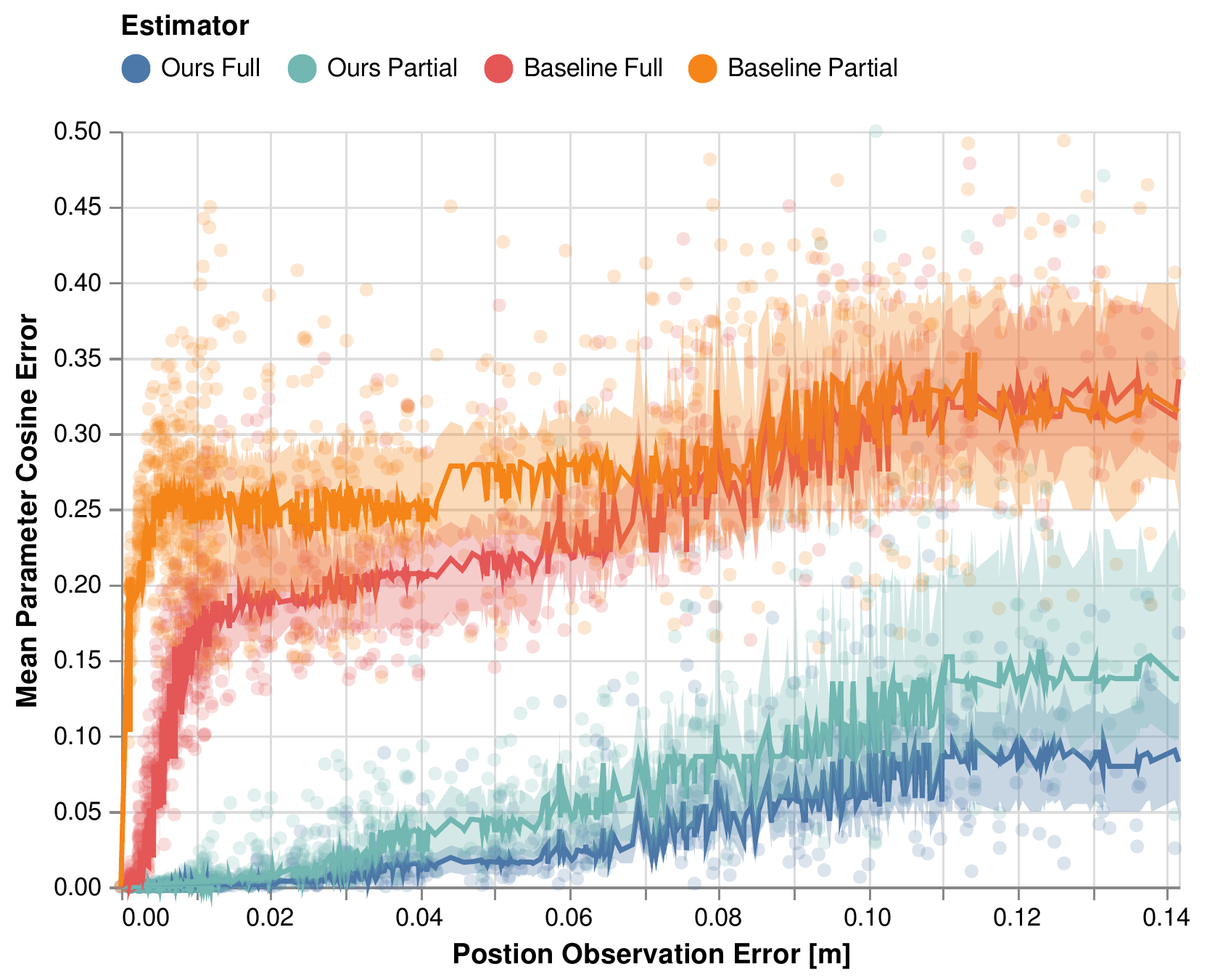}
    }
    \hfill
    \subfigure[Trajectory reconstruction\label{fig:runexmp-estimator_position_error}]{
    \includegraphics[scale=\vegascale]{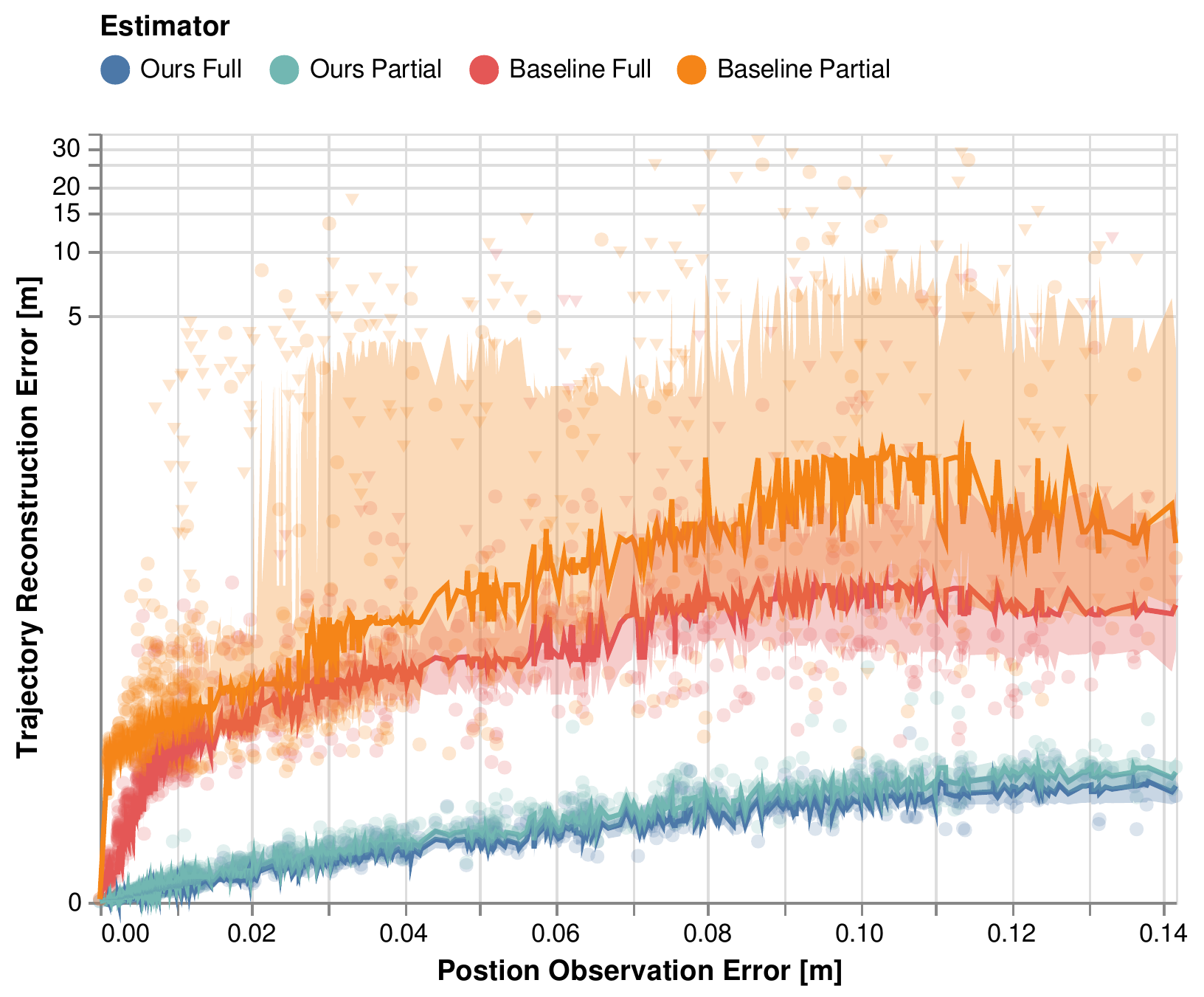}
    }
    \caption{Estimation performance of our method and the baseline for the 2-player collision-avoidance example, with noisy full and partial state observations.
    (a) Error measured directly in parameter space using \cref{eqn:cosine_metric}.
    (b) Error measured in position space using \cref{eqn:reconstruction_metric}.
    Triangular data markers in (b) highlight objective estimates which lead to ill-conditioned games.
    Solid lines and ribbons indicate the median and \ac{iqr} of the error for each case.
    }
    \label{fig:runexmp-estimator_statistics}
\end{figure}

\Cref{fig:runexmp-estimator_statistics} shows the estimator performance for varying levels of observation noise in two different metrics.
\cref{fig:runexmp-estimator_parameter_error} reports the mean cosine error of the objective parameter estimates.
That is, we measure cosine-dissimilarity between the unobserved true model parameters~$\costparams{}_\textrm{true}$ and the learned estimates~$\costparams{}_\textrm{est}$ according to
\begin{align}\label{eqn:cosine_metric}
     D_\textrm{cos}(\costparams{}_\textrm{true}, \costparams{}_\textrm{est}) = 1 -
     \frac{1}{\numplayers} \sum_{i \in [\numplayers]} \frac{\costparams{i\top}_\textrm{true} \costparams{i}_\textrm{est}}{\norm{\costparams{i}_\textrm{true}}_2 \norm{\costparams{i}_\textrm{est}}_2},
\end{align}
where the mean is taken over the~$\numplayers$ players.
The normalization of the parameter vectors in \cref{eqn:cosine_metric} reflects the fact that the absolute scaling of each player's objective parameters does not effect their optimal behavior, holding other players' parameters fixed.
In sum, this metric measures the estimator performance in objective \emph{parameter space}.

\cref{fig:runexmp-estimator_position_error} shows the mean absolute position error for trajectory \emph{reconstructions} computed by finding a root of \cref{eqn:forward_kkt_conditions} using the estimated objective parameters.
Reconstruction error allows us to inspect the quality of learned cost parameters for explaining observed vehicle motion, providing a more tangible metric of algorithmic quality.
In addition to the raw data, we highlight the median as well as the \ac{iqr} of the estimation error over a rolling window of 60 data points.

\Cref{fig:runexmp-estimator_parameter_error} shows that both our method and the baseline recover the true parameters~$\costparams{}$ reliably even for partial observations, if the observations are noiseless.
However, the performance of the baseline degrades rapidly with increasing noise variance.
This pattern is particularly pronounced in the setting of partial observations.
On the other hand, our estimator recovers the unknown cost parameters more accurately in both settings, and with a smaller variance than the baseline.
Thus, compared to the \ac{kkt} residual baseline, the performance of our method degrades gracefully when both full and partial observations are corrupted by noise.

Next, we study these methods' relative performance as measured by reconstruction error, as shown in \cref{fig:runexmp-estimator_position_error}.
Here, reconstruction error is measured according to
\begin{align}\label{eqn:reconstruction_metric}
     D_\textrm{rec}(\costparams{}_\textrm{true}, \costparams{}_\textrm{est}) =
     \frac{1}{\numplayers \horizon} \sum_{i \in [\numplayers]} \sum_{t \in [\horizon]} \|p^i_{\textrm{rec}, t} - p^i_{\textrm{true}, t}\|_2,
\end{align}
where~$p^i_{\textrm{true}, t}$ denotes the true position of \player{i} at time step~$t$, and~$p^i_{\textrm{rec}, t}$ denotes the position reconstructed from a Nash solution to the game with estimated cost parameters~$\costparams{}_\textrm{est}$.
We see similar patterns here as in the parameter error space, indicating the reliability of our method in both noisy full and partial observation settings.

Additionally, note that we have denoted some data points for the baseline method with triangular markers.
For these Monte Carlo samples, the learned parameters~$\costparams{}_\textrm{est}$ specify ill-conditioned objectives that prevent us from recovering roots of \cref{eqn:forward_kkt_conditions}---essentially rendering the parameter estimates useless for downstream applications.
This can happen, for example, when proximity costs dominate control input costs.
For the baseline, a total of 104 out of 880 estimates result in an ill-conditioned forward game when states are fully observed.
In the case of partial observations, the number of learning failures increases to 218.
In contrast, our method recovers well-conditioned player objectives for all demonstrations and allows for accurate reconstruction of the game trajectory.

\begin{figure}
  \centering
  \subfigure[Demonstrations]{
  \includegraphics[scale=\vegascale]{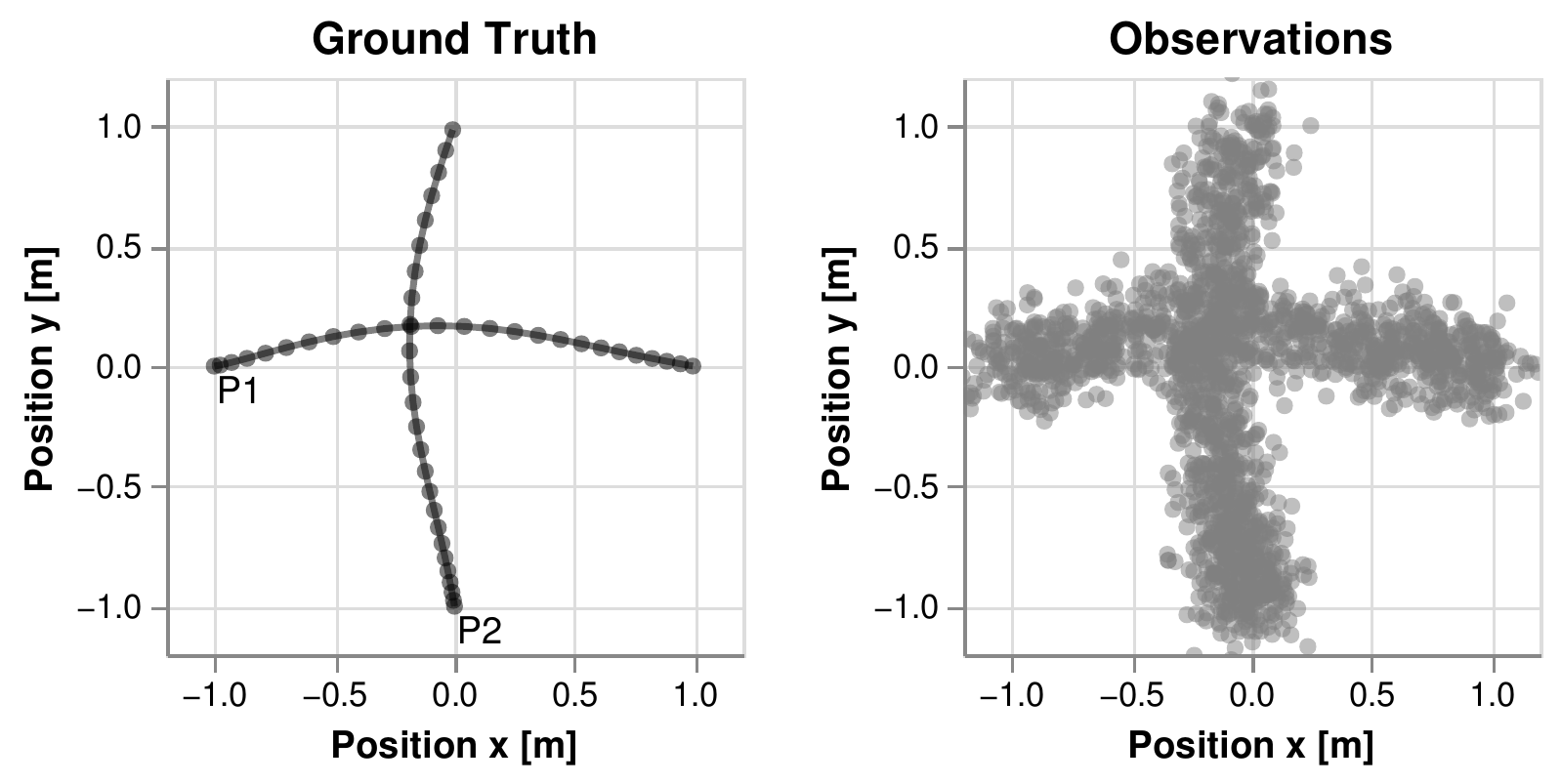}
  \label{fig:runexmp-obs_trajs}
  }
  \hfill
  \subfigure[Ours]{
  \includegraphics[scale=\vegascale]{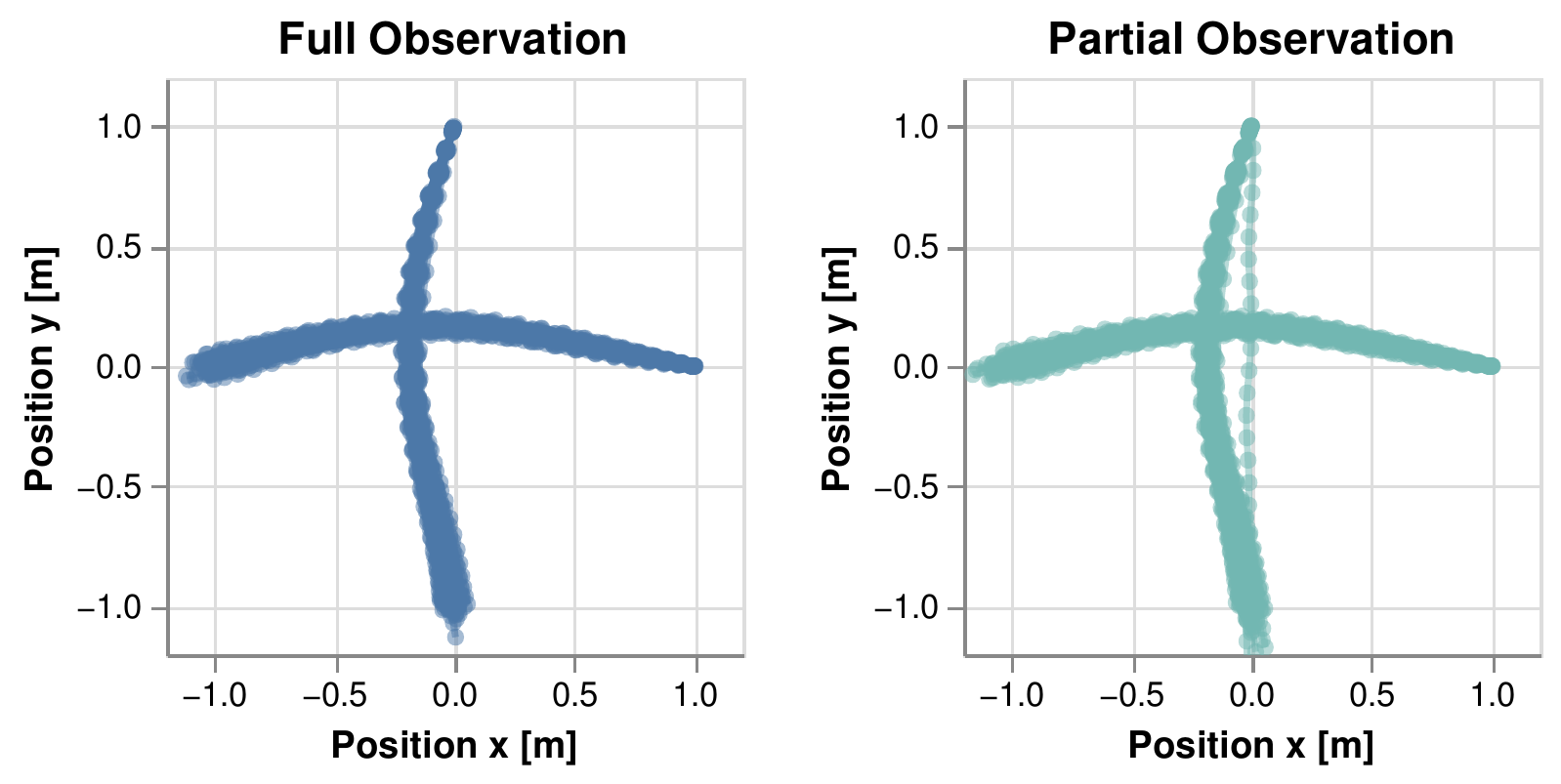}
  \label{fig:runexmp-our_trajs}
  }
  \hfill
  \subfigure[Baseline]{
  \includegraphics[scale=\vegascale]{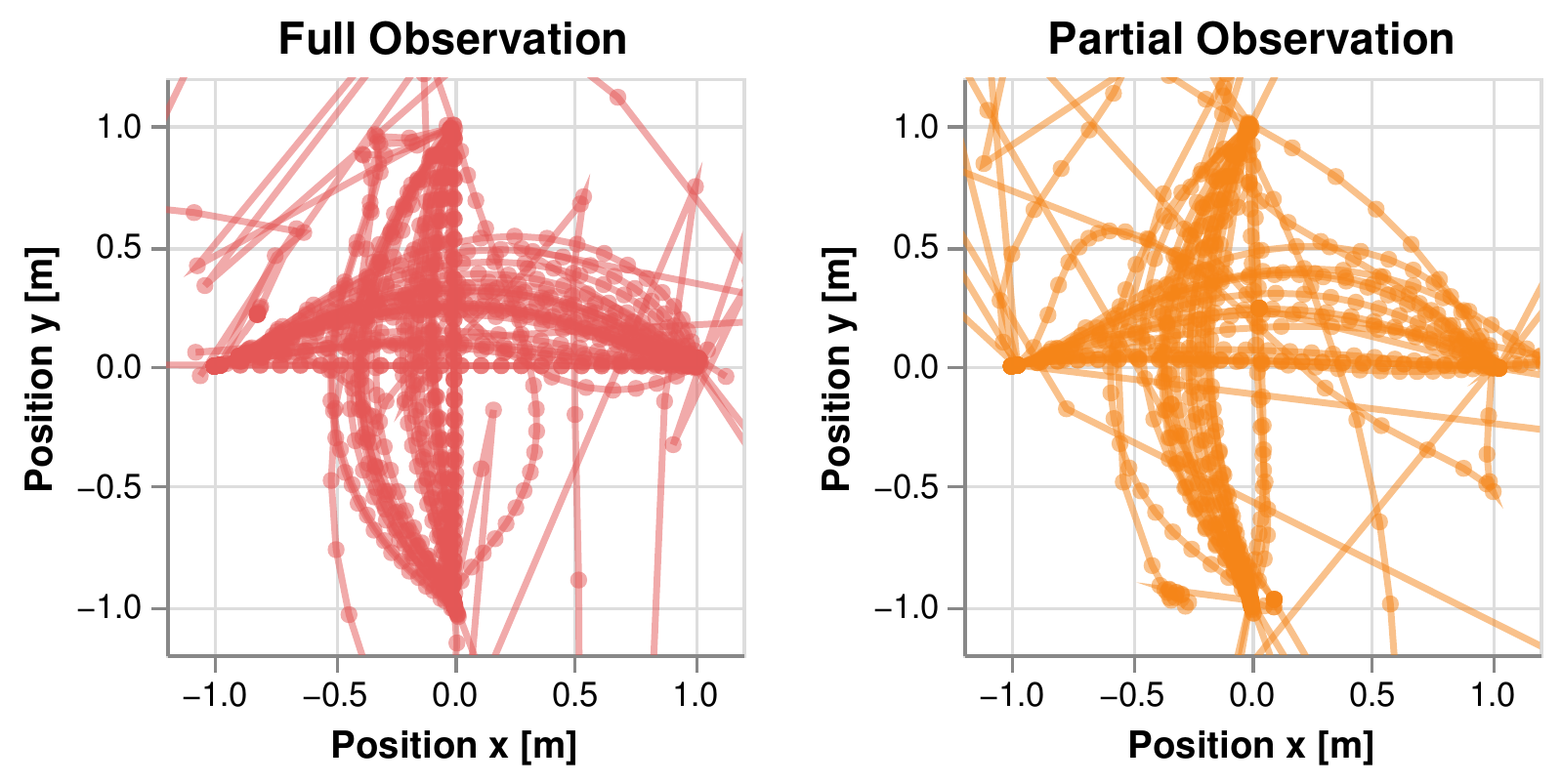}
  \label{fig:runexmp-baseline_trajs}
  }
  \caption{Qualitative reconstruction performance for the 2-player collision avoidance example at noise level~$\sigma = 0.1$ for 40 different observation sequences. (a) Ground truth trajectory and observations, where each player wishes to reach a goal location opposite their initial position. (b, c) Trajectories recovered by solving the game at the estimated parameters for our method and the baseline using noisy full and partial state observations.
h  }
  \label{fig:runexmp-trajs}
\end{figure}

For additional intuition of the performance gap, \cref{fig:runexmp-trajs} visualizes the reconstruction results in trajectory space for a fixed initial condition.
\Cref{fig:runexmp-obs_trajs} shows the noise corrupted demonstrations generated for isotropic \ac{awgn} with standard deviation~$\sigma = 0.1$.
\Cref{fig:runexmp-our_trajs} and \cref{fig:runexmp-baseline_trajs} show the corresponding trajectories reconstructed by solving the game using the objective parameters learned by our method and the baseline, respectively.
Note that our method generates a far smaller fraction of outliers than the baseline.
Furthermore, the performance of our method is only marginally effected by partial state observability, whereas baseline performance degrades substantially.

\subsubsection{Online Learning with Prediction}
\label{sec:detailed_offline_prediction_experiments}
Next, we study the performance of both our proposed method and the \ac{kkt} residual baseline in the setting of objective learning with prediction.
Following the problem description of \cref{sec:offline_learning_prediction}, here, only the beginning of an unfolding dynamic game is observed.
This problem naturally describes a single time frame of online operations where observations accumulate as time evolves.

We conduct a Monte Carlo analysis of the two-player collision-avoidance game from \cref{sec:experimental_setup} in which we vary the number of observed time steps of a fixed-length game.
For this truncated observation sequence, each method is tasked to learn the players' underlying cost parameters~$\costparams{i}$ \emph{and} predict their motion for the next~$s_\mathrm{p} = 10$ time steps.
Our method accomplishes these coupled tasks jointly by solving \cref{eqn:inverse_approach_prediction_lsq}.
The \ac{kkt} residual baseline, however, operates on the estimates provided the preceding smoothing step, therefore, cannot couple unobserved, future time steps with cost inference.
Instead, it achieves this task in a two-stage procedure: First, parameter estimates are recovered from a truncated game over only the observed~$\obshorizon$ time steps.
With these parameters in hand, the baseline then predicts future game states by re-solving a forward game starting from the final state estimate~$\tilde{\state}_\obshorizon$ with time steps simulated from~$t \in \{\obshorizon, \dots, \obshorizon + 10\}$.

\begin{figure}
  \centering
  \subfigure[]{
    \includegraphics[scale=\vegascale]{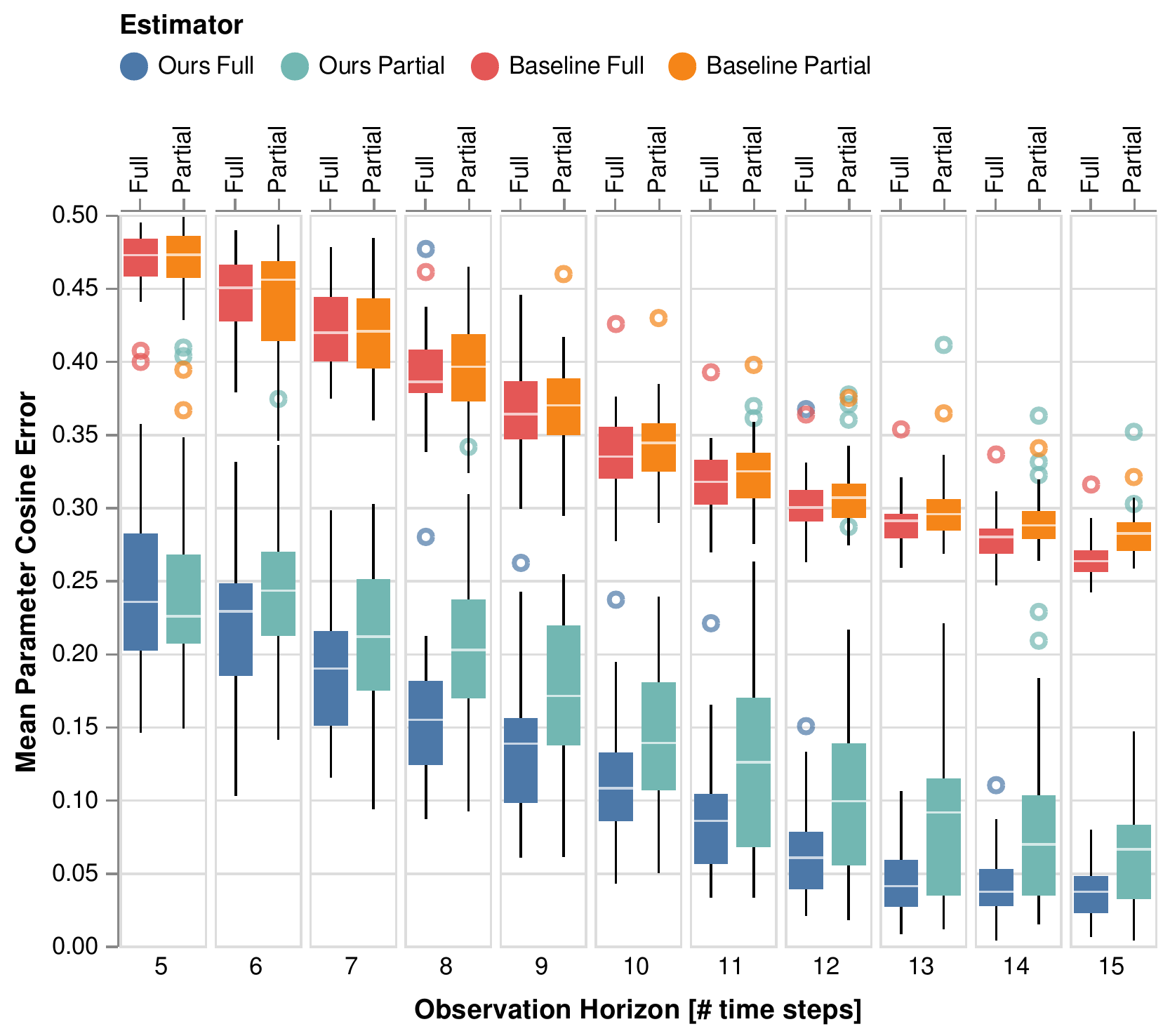}
    \label{fig:2player_prediction_param_error}
  }
  \hspace{-5pt}
  \subfigure[]{
    \includegraphics[scale=\vegascale]{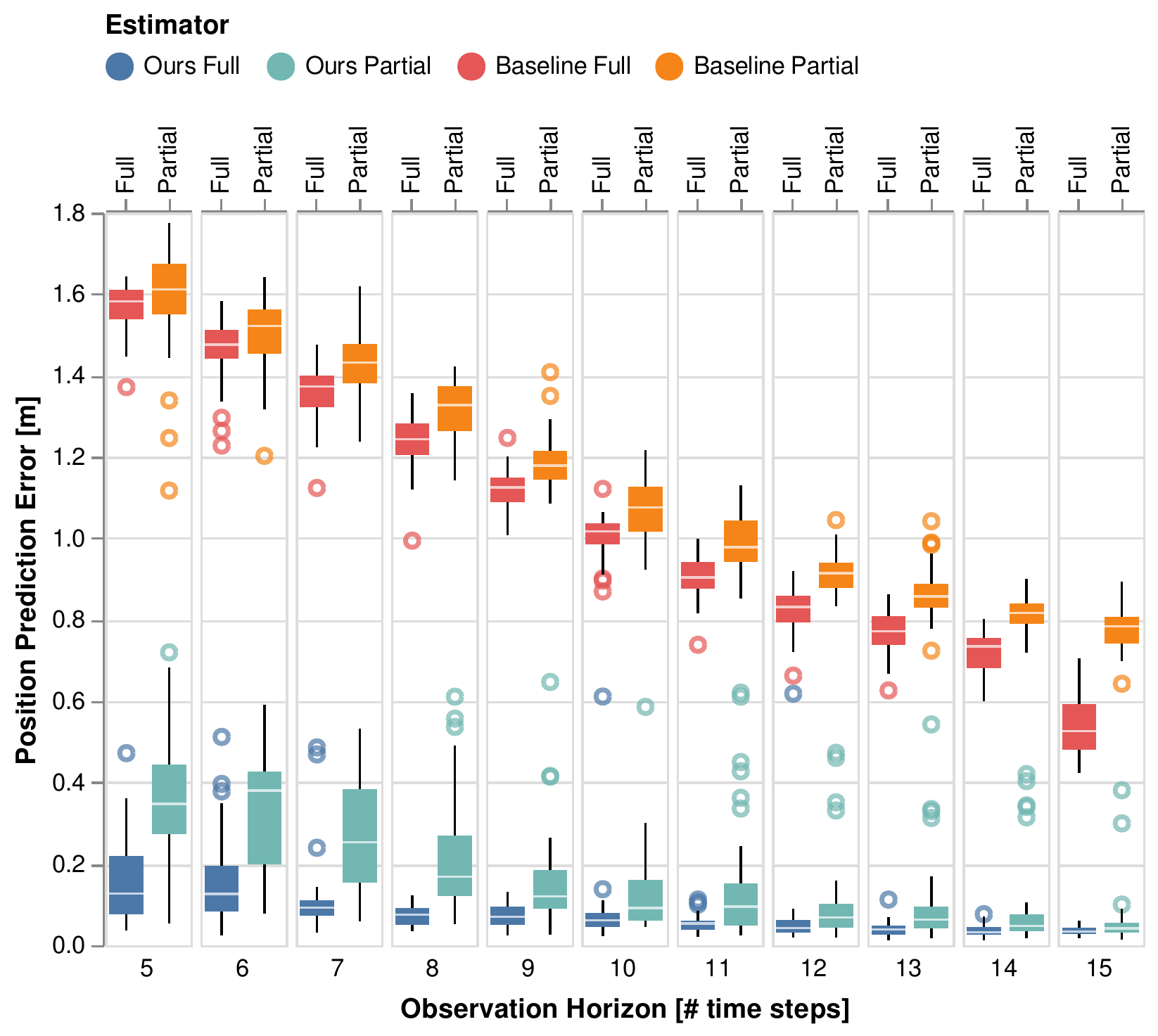}
    \label{fig:2player_prediction_prediction_error}
  }
  \caption{Estimation performance for our method and the baseline for varying numbers of observations of the 2-player collision-avoidance example at a fixed noise level of~$\sigma = 0.05$. (a) Estimation performance measured directly in parameter space using \cref{eqn:cosine_metric}. (b) Prediction error over the next \SI{10}{\second} beyond the observation horizon using \cref{eqn:reconstruction_metric}.
  }
  \label{fig:2player_online_prediction}
\end{figure}

In \cref{fig:2player_online_prediction}, we vary the observation horizon \mbox{$\obshorizon \in \{5, \dots, 15\}$} for a ground-truth game played over~$25$ time steps.
For each value of~$\obshorizon$, we sample 40 sequences of observations~$\{\observation_t\}_{t=1}^\obshorizon$.
Here, we fix an isotropic Gaussian noise level of~$\sigma = 0.05$, and measure the performance of both our method and the baseline using two distinct metrics.
In \cref{fig:2player_prediction_param_error}, we measure learning performance in parameter space using the metric given in \cref{eqn:cosine_metric}.
As shown, our approach consistently estimates the cost parameters more accurately than the baseline.
Furthermore, as the observation horizon~$\obshorizon$ increases, both methods improve. 
In \cref{fig:2player_prediction_prediction_error}, we see that these patterns persist when we measure performance in trajectory space, applying the metric of \cref{eqn:reconstruction_metric} to the predicted states~$\state_t, t \in \{\obshorizon, \dots, \obshorizon + 10\}$.
Indeed, in this case, the performance gap is even more pronounced.
By observing only~$\obshorizon=5$ steps, our method reliably outperforms the baseline even when the baseline is given triple the number of observations.

\begin{figure}
    \centering
    \includegraphics[scale=\vegascale]{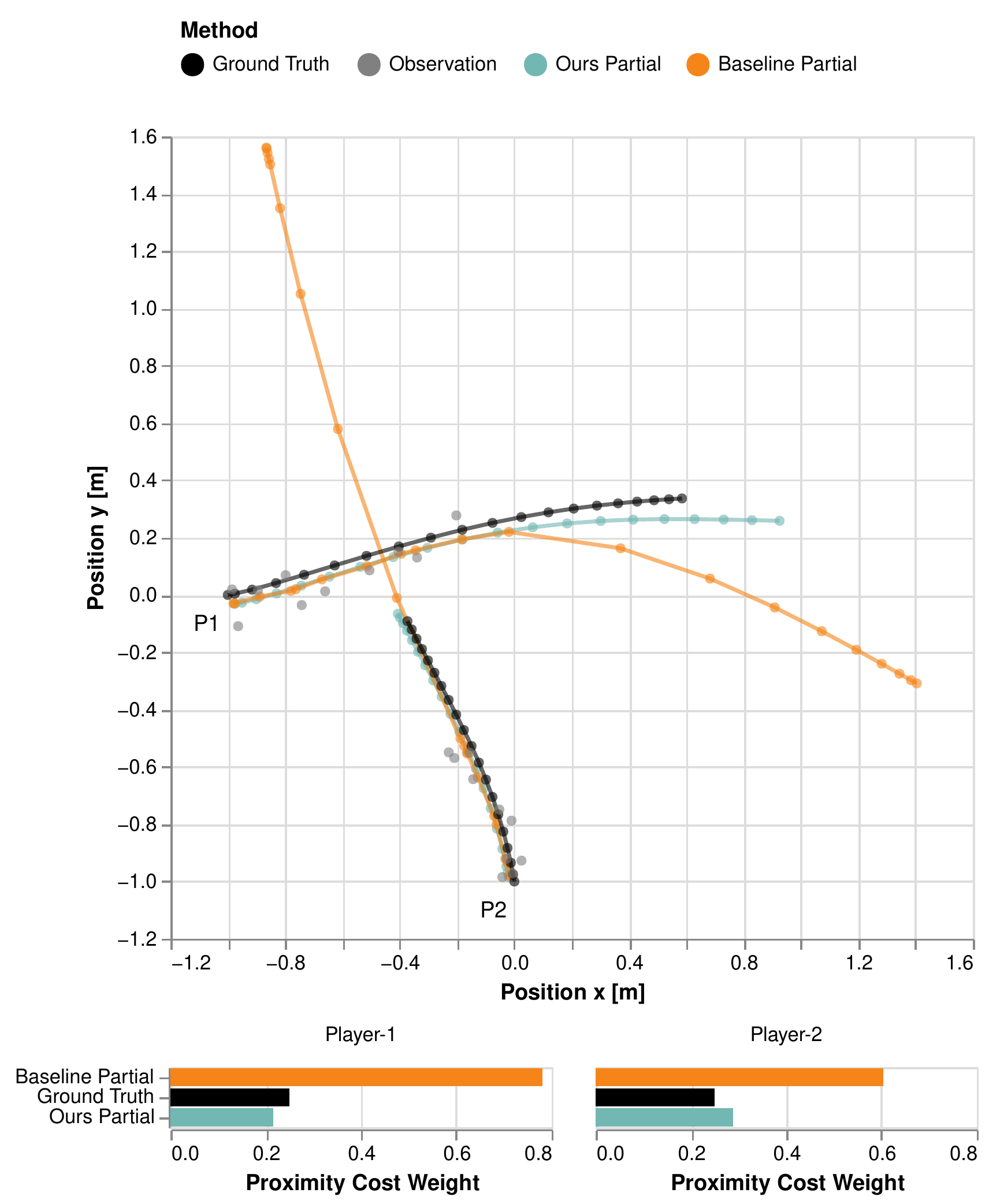}
    \caption{
    Qualitative prediction performance of our method and the baseline for the 2-player collision avoidance example when only the first 10 out of 25 time steps are observed.
    }
    \label{fig:2player_online_prediction_trajectories}
\end{figure}

To inspect these results more closely, in \cref{fig:2player_online_prediction_trajectories} we show the output of both methods for a single observation sequence of length~$\obshorizon = 10$.
This visualization highlights a key advantage of our approach compared with the baseline.
In this scenario, \player{2} (bottom) turns left early on in order to avoid \player{1} (left) later along the path to its goal.
Their ground truth trajectories are shown in black.
However, the methods only receive noise-corrupted partial state observations of the first~$\obshorizon = 10$ time steps shown in gray.
Our method models the players' interactions as continuing into the future, allowing it to attribute observed behavior to future costs.
In this instance, our method correctly explains \player{2}'s observed left turn as the result of a modest penalty on proximity, which becomes important only later in the trajectory when the players are close to one another.
Cost estimation is shown at the bottom of \cref{fig:2player_online_prediction_trajectories}. 
The \ac{kkt} residual baseline is incapable of such attributions.
More precisely, it can only consider the \ac{kkt} residuals~$\kktresidual(\cdot; \costparams{})$ of \cref{eqn:kkt_residual_baseline} for time steps~$t \in [\obshorizon]$.
Hence, the baseline must presume that the game terminates at~$\obshorizon$ rather than at some time in the future.
Thus, it cannot anticipate the immediate future consequences of particular cost models.
In \cref{fig:2player_online_prediction_trajectories}, the baseline can only explain the players' early observed collision avoidance maneuver with an extremely large penalty on proximity to their opponents.
As a result, it predicts that the players will quickly drive away from one another.
Unlike our method, the baseline's prediction rapidly diverges from the ground truth.

\begin{figure}
    \centering
    \includegraphics[scale=\vegascale]{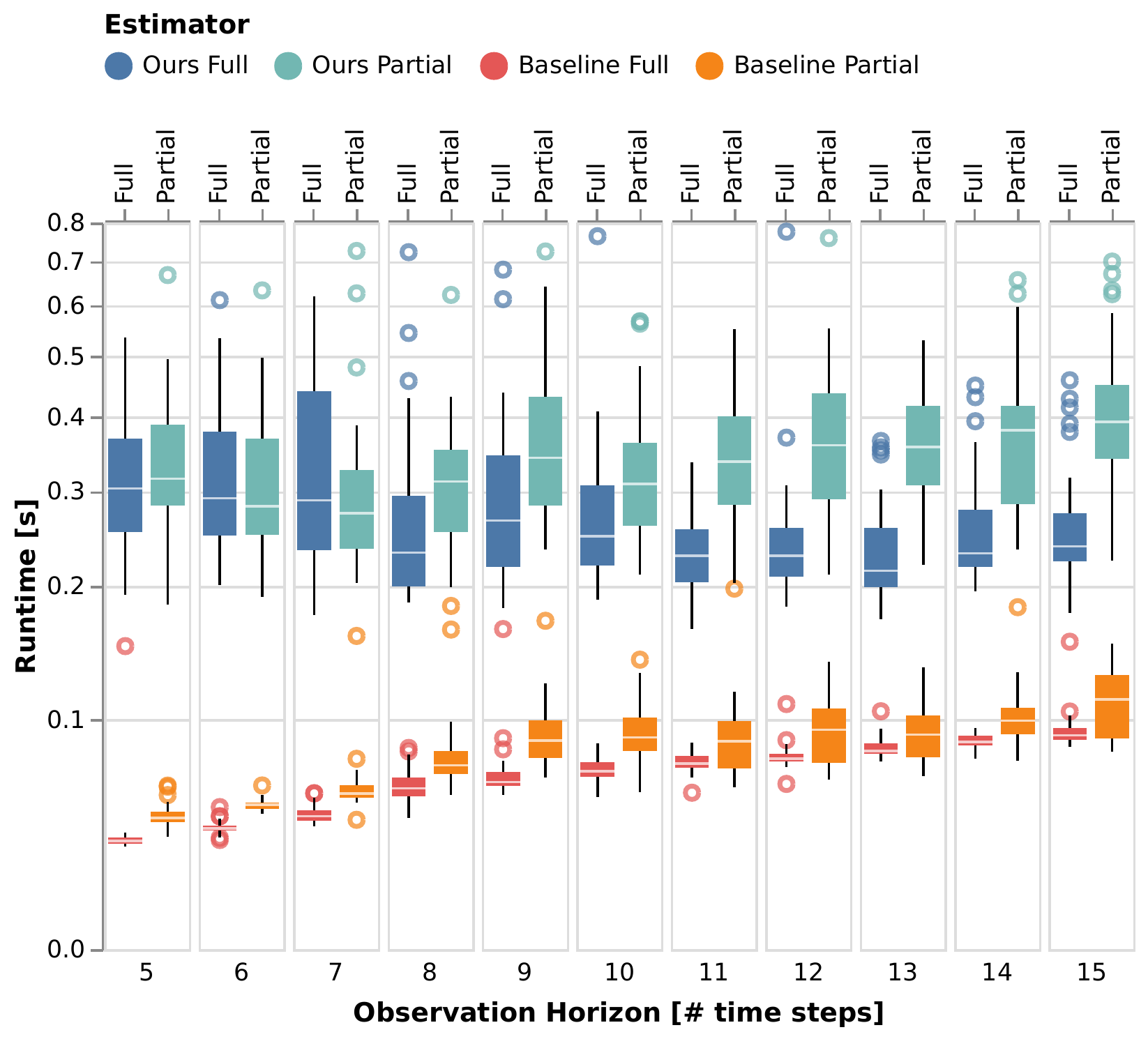}
    \caption{
    \revised{
    Runtime of our method and the baseline for varying numbers of observations of the 2-player collision-avoidance example at a fixed noise level of $\sigma = 0.05$.
    }
    }
    \label{fig:2player_online_timing}
\end{figure}

\revised{
Beyond inference and prediction accuracy, a key factor for online operation is the computational complexity.
To investigate this point,~\cref{fig:2player_online_timing} shows the computation time of both methods for the same dataset underpinning~\cref{fig:2player_online_prediction}.
These timing results were obtained on a AMD Ryzen 9 5900HX laptop CPU.
Overall, we observe that the \ac{kkt} residual baseline has a lower runtime than our approach.
The reduced runtime can be  attributed to the fact that, by fixing the states and inputs a priori, the \ac{kkt} residual formulation yields a simpler \emph{convex} optimization problem in~\cref{eqn:kkt_residual_baseline}.
Nonetheless, our method's runtime still remains moderate and scales gracefully with the observation horizon.
We note that our current implementation is not optimized for speed.
In practical applications in the context of receding-horizon applications---a topic that we shall discuss in~\cref{sec:online_learning_experiments}---the runtime may be further reduced via improved warm-starting and memory sharing across planner invocations.
}

\subsection{Scaling to Larger Games}
\label{sec:scaling_experiments}

While our approach is more easily analyzed in the small, two-player collision-avoidance game of \cref{sec:detailed_experiments}, it readily extends to larger multi-agent interactions.
In order to demonstrate scalability of the approach, we therefore replicate the offline learning analysis of \cref{sec:detailed_offline_experiments} in a larger 5-player highway driving scenario depicted in \cref{fig:frontfig}.
Finally, we demonstrate a proof of concept for online, receding horizon learning in this scaled setting following the setup of \cref{sec:online_learning}.

In the highway scenario discussed through the remainder of this section, each player wishes to make forward progress in a particular lane at an unknown nominal speed, rather than reach a desired position as above.
Therefore, ground-truth objectives use a quadratic penalty on deviation from a desired state that encodes each player's target lane and preferred travel speed rather than a specific goal location.
Despite these differences, this class of objectives is still captured by the cost structure introduced in \cref{eqn:runexp-cost}.

\subsubsection{Offline Learning}
\label{sec:scaling_offline_experiments}

\begin{figure}
    \centering
    \subfigure[Parameter estimation\label{fig:highway-estimator_parameter_error}]{
    \includegraphics[scale=\vegascale]{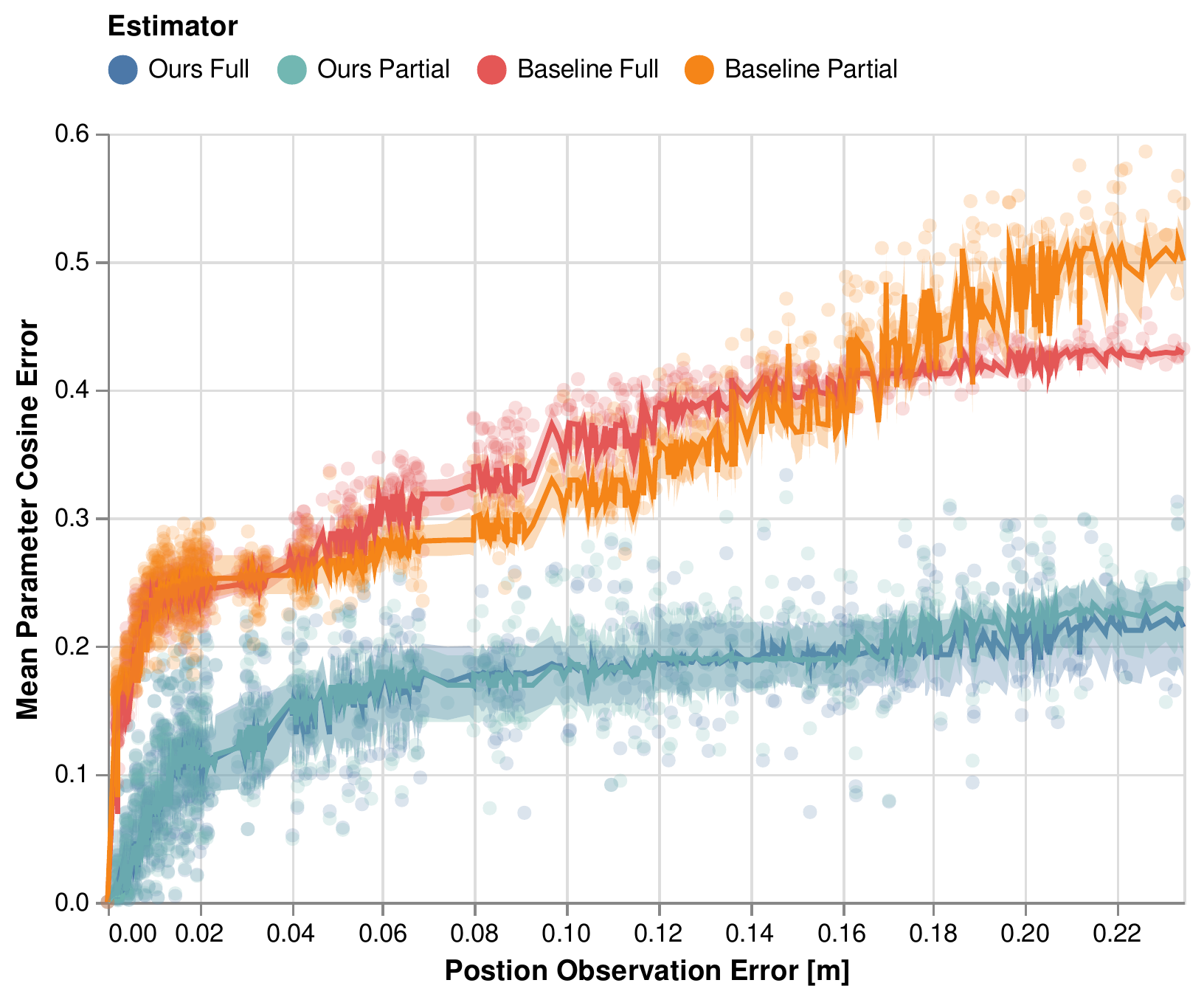}
    }
    \hfill
    \subfigure[Trajectory Reconstruction\label{fig:highway-estimator_position_error}]{
    \includegraphics[scale=\vegascale]{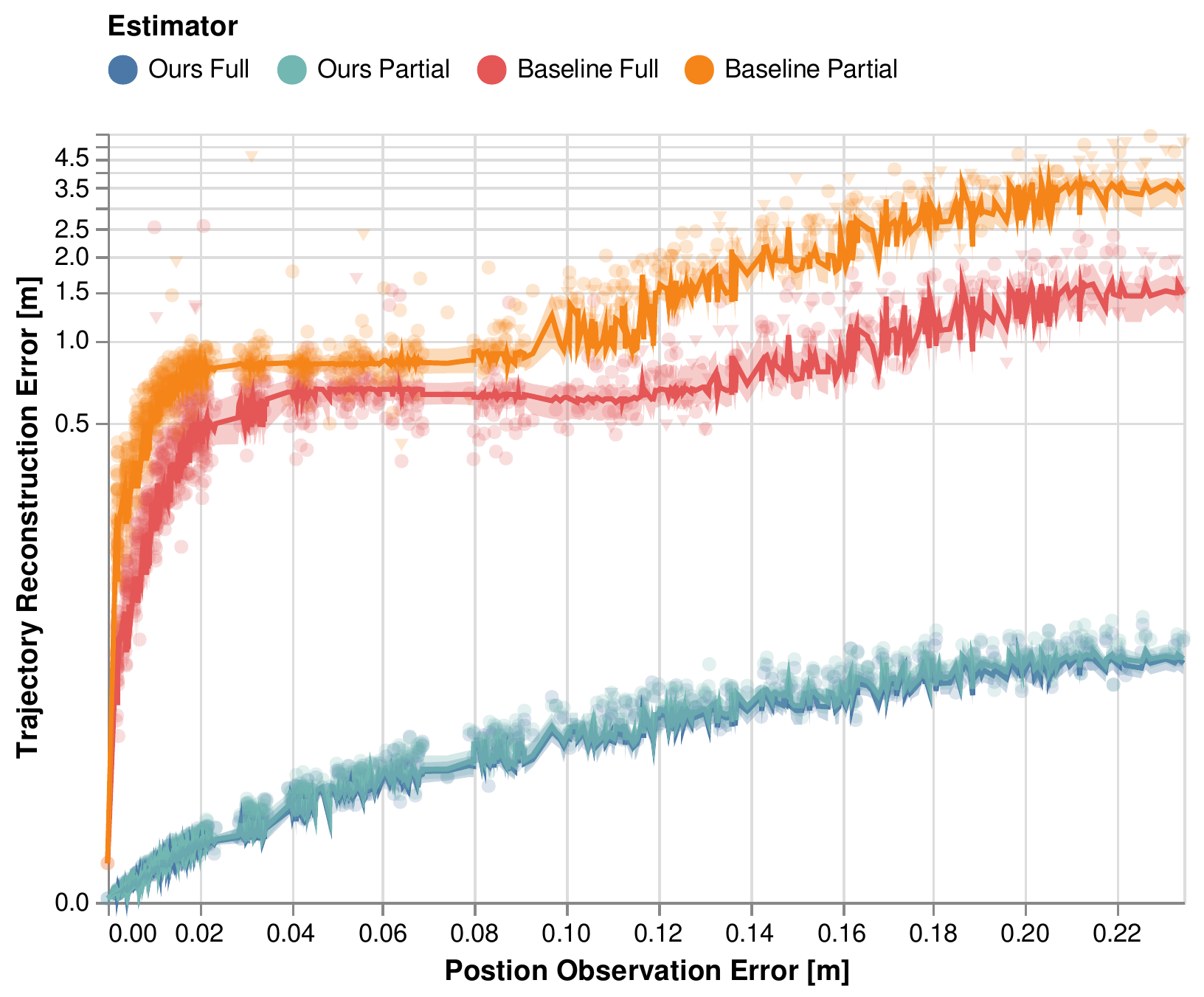}
    }
    \caption{Estimation performance of our method and the baseline for the 5-player highway overtaking example, with noisy full and partial state observations.
    (a) Error measured directly in parameter space using \cref{eqn:cosine_metric}.
    (b) Error measured in position space using \cref{eqn:reconstruction_metric}.
    Triangular data markers in (b) highlight objective estimates which lead to ill-conditioned games.
    Solid lines and ribbons indicate the median and \ac{iqr} of the error for each case.
    \label{fig:highway-estimator_statistics}
    }
\end{figure}

First, we study the performance of our method and the \ac{kkt} residual baseline in the setting of offline learning without trajectory prediction.
\Cref{fig:highway-estimator_statistics} displays these results, using the same metrics as in \cref{sec:detailed_offline_experiments} to measure performance in parameter space---\cref{fig:highway-estimator_parameter_error}---and position space---\cref{fig:highway-estimator_position_error}.
As before, our method demonstrably outperforms the baseline in both fully and partially observed settings.
Furthermore, whereas our method performs comparably according to both metrics in the full and partial observation settings, the baseline performance differs between the two metrics.
That is, while the performance of the baseline measured in parameter space is not significantly effected by less informative observations, the effect is significant in trajectory space.
This inconsistency can be attributed the fact that certain objective parameters have stronger influence on the resulting game trajectory than others. 
Since our method's objective is observation fidelity, here measured by the measurement likelihood of \cref{eqn:inverse_approach_objective}, it directly accounts for these varying sensitivities.
The baseline, however, greedily optimizes the \ac{kkt} residual of \cref{eqn:kkt_residual_baseline}, irrespective of the resulting equilibrium trajectory.

\subsubsection{Online Learning and Receding Horizon Prediction}
\label{sec:online_learning_experiments}

Finally, we demonstrate the application of our method for simultaneous online learning and receding-horizon prediction in the \mbox{5-player} highway navigation scenario depicted in \cref{fig:frontfig}.

\begin{figure}
  \centering
    \hspace{-8pt}
    \includegraphics[scale=\vegascale]{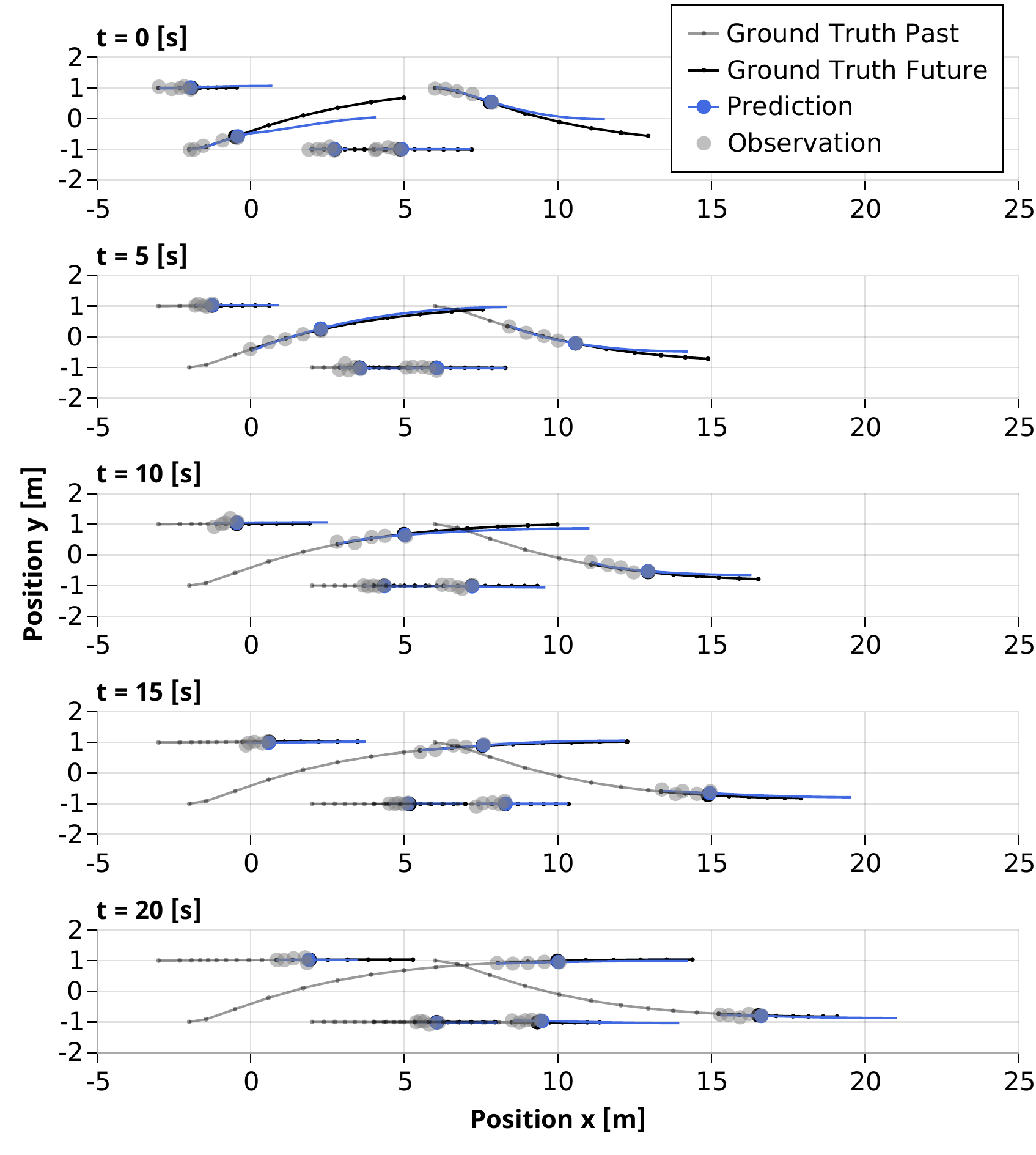}
  \caption{Demonstration of our method in an \emph{online} application of simultaneous objective learning and trajectory prediction for the 5-player highway navigation scenario. At each time step, objective learning is performed on a fixed-lag buffer of \SI{5}{\second} of observation data which is coupled with trajectory prediction \SI{10}{\second} into the future.
  }
  \label{fig:receding_horizon_experiments}
\end{figure}

Here, the information available to the estimator evolves over time and the problem only admits access to \emph{past} observations of the game state for cost learning.
Following the proposed procedure of \cref{sec:online_learning}, here, we limit the computational complexity of the estimation problem by considering only a fixed-lag buffer of observations over the last $\si{5}{s}$ and predict all player's behavior over the next $\si{10}{s}$.
The qualitative performance of our method under noise-corrupted partial state observation is shown in \cref{fig:receding_horizon_experiments}.
As can be seen, from only a few seconds of data, our method learns player objectives that accurately predict the evolution of the game over a receding prediction horizon.
Note that, by design, objective learning and behavior prediction is achieved \emph{simultaneously} by solving a single joint optimization problem as in \cref{eqn:inverse_approach}.
This ability to couple online learning and prediction makes it particularly suitable for online applications.

\section{Conclusion}
\label{sec:conclusion}

In this paper, we have introduced a novel approach to learn the parameters of players' objectives in dynamic, noncooperative interactions, given only noisy, partial observations.
This \emph{inverse} dynamic game arises in a wide variety of multi-robot and human-robot interactions and generalizes well-studied problems such as \acl{ioc}, \acl{irl}, and learning from demonstrations.
Contrary to prior work, our method learns players' cost parameters while \emph{simultaneously} recovering the forward game trajectory consistent with those parameters, with overall performance measured according to observation fidelity.
We have shown how this formulation naturally extends to both \emph{offline} learning and prediction problems, as well as \emph{online}, receding horizon learning.

We have conducted extensive numerical simulations to characterize the performance of our method and compare it to a state-of-the-art baseline method \citep{rothfuss2017ifac,awasthi2020acc}.
These simulations clearly demonstrate our method's improved robustness to both observation noise and partial observations.
Indeed, existing methods presume noiseless, full state observations and thus require \emph{a priori} estimation of states and inputs.
Our method recovers objective parameters, reconstructs past game trajectories, and predicts future trajectories far more accurately than the baseline.
Beyond that, our method's structure allows to perform all of these tasks jointly as the solution of a single optimization problem.
This feature renders our method suitable for online learning and prediction in a receding horizon fashion.

In light of these encouraging results, there are several directions for future research.
Most immediately, our method lends itself naturally to deployment onboard physical robotic systems such as the autonomous vehicles considered in the examples of \cref{sec:experiments}.
In particular, the online, receding horizon learning and prediction procedure of \cref{sec:online_learning} may be run onboard an autonomous car.
Here, the ``ego'' agent would seek to learn other vehicles' objective parameters while simultaneously using the receding horizon game solution to respond to predicted opponent strategies.

Another exciting, more theoretical direction consists of extending our formulation to more complex equilibrium concepts than \ac{olne}.
For example, recent solution methods for forward games in state feedback Nash equilibria \citep{fridovich2020icra, laine2021arxiv, di2021newton} might be adapted to solve inverse games along the lines of \cref{eqn:inverse_problem}.

\section{Declaration of Conflicts of Interest}
The authors declare that there is no conflict of interest.

\bibliographystyle{SageH}
\bibliography{glorified,new}

\begin{thebibliography}{42}
\providecommand{\natexlab}[1]{#1}
\providecommand{\url}[1]{\texttt{#1}}
\providecommand{\urlprefix}{URL }
\expandafter\ifx\csname urlstyle\endcsname\relax
  \providecommand{\doi}[1]{DOI:\discretionary{}{}{}#1}\else
  \providecommand{\doi}{DOI:\discretionary{}{}{}\begingroup
  \urlstyle{rm}\Url}\fi

\bibitem[{Albrecht et~al.(2011)Albrecht, Ramirez-Amaro, Ruiz-Ugalde,
  Weikersdorfer, Leibold, Ulbrich and Beetz}]{albrecht2011humanoids}
Albrecht S, Ramirez-Amaro K, Ruiz-Ugalde F, Weikersdorfer D, Leibold M, Ulbrich
  M and Beetz M (2011) Imitating human reaching motions using physically
  inspired optimization principles.
\newblock In: \emph{Proc.~of the IEEE Intl.~Conf.~on Humanoid Robots}. IEEE.

\bibitem[{Andersson et~al.(2019)Andersson, Gillis, Horn, Rawlings and
  Diehl}]{andersson2019mpc}
Andersson JAE, Gillis J, Horn G, Rawlings JB and Diehl M (2019) {CasADi}: a
  software framework for nonlinear optimization and optimal control.
\newblock \emph{Mathematical Programming Computation} 11(1): 1--36.

\bibitem[{Awasthi(2019)}]{awasthi2019master}
Awasthi C (2019) \emph{Forward and Inverse Methods in Optimal Control and
  Dynamic Game Theory}.
\newblock Master's Thesis, University of Minnesota.

\bibitem[{Awasthi and Lamperski(2020)}]{awasthi2020acc}
Awasthi C and Lamperski A (2020) Inverse differential games with mixed
  inequality constraints.
\newblock In: \emph{Proc.~of the IEEE American Control Conference (ACC)}. IEEE.

\bibitem[{Ba{\c{s}}ar and Olsder(1999)}]{basar1998gametheorybook}
Ba{\c{s}}ar T and Olsder GJ (1999) \emph{Dynamic noncooperative game theory},
  volume~23.
\newblock Society for Industrial and Applied Mathematics (SIAM).

\bibitem[{Bezanson et~al.(2017)Bezanson, Edelman, Karpinski and
  Shah}]{bezanson2017sirev}
Bezanson J, Edelman A, Karpinski S and Shah VB (2017) {J}ulia: A fresh approach
  to numerical computing.
\newblock \emph{SIAM Review (SIREV)} 59(1): 65--98.

\bibitem[{Daskalakis et~al.(2009)Daskalakis, Goldberg and
  Papadimitriou}]{daskalakis2009complexity}
Daskalakis C, Goldberg PW and Papadimitriou CH (2009) The complexity of
  computing a {N}ash equilibrium.
\newblock \emph{SIAM Journal on Computing} 39(1): 195--259.

\bibitem[{Di and Lamperski(2019)}]{di2019cdc}
Di B and Lamperski A (2019) Newton’s method and differential dynamic
  programming for unconstrained nonlinear dynamic games.
\newblock In: \emph{Proceedings of the Conference on Decision Making and
  Control (CDC)}. IEEE.

\bibitem[{Di and Lamperski(2021)}]{di2021newton}
Di B and Lamperski A (2021) {N}ewton’s method, {B}ellman recursion and
  differential dynamic programming for unconstrained nonlinear dynamic games.
\newblock \emph{Dynamic Games and Applications} : 1--49.

\bibitem[{Dirkse and Ferris(1995)}]{dirkse1995path}
Dirkse SP and Ferris MC (1995) The path solver: a nommonotone stabilization
  scheme for mixed complementarity problems.
\newblock \emph{Optimization methods and software} 5(2): 123--156.

\bibitem[{Dunning et~al.(2017)Dunning, Huchette and Lubin}]{dunning2017sirev}
Dunning I, Huchette J and Lubin M (2017) {JuMP}: A modeling language for
  mathematical optimization.
\newblock \emph{SIAM Review (SIREV)} 59(2): 295--320.

\bibitem[{Englert and Toussaint(2018)}]{englert2018ijrr}
Englert P and Toussaint M (2018) Inverse {KKT}: Learning cost functions of
  manipulation tasks from demonstrations.
\newblock \emph{Intl.~Journal~of Robotics Research (IJRR)} : 57--72.

\bibitem[{Ferris et~al.(2005)Ferris, Dirkse and
  Meeraus}]{ferris2005mathematical}
Ferris MC, Dirkse SP and Meeraus A (2005) Mathematical programs with
  equilibrium constraints: Automatic reformulation and solution via constrained
  optimization.
\newblock \emph{Frontiers in applied general equilibrium modeling} : 67--93.

\bibitem[{Fridovich-Keil et~al.(2020)Fridovich-Keil, Ratner, Peters, Dragan and
  Tomlin}]{fridovich2020icra}
Fridovich-Keil D, Ratner E, Peters L, Dragan AD and Tomlin CJ (2020) Efficient
  iterative linear-quadratic approximations for nonlinear multi-player
  general-sum differential games.
\newblock In: \emph{Proc.~of the IEEE Intl.~Conf.~on Robotics \& Automation
  (ICRA)}. IEEE.

\bibitem[{Gallager(2013)}]{gallager2013stochastic}
Gallager RG (2013) \emph{Stochastic processes: theory for applications}.
\newblock Cambridge University Press.

\bibitem[{Gill et~al.(2005)Gill, Murray and Saunders}]{gill2005sirev}
Gill PE, Murray W and Saunders MA (2005) {SNOPT}: An {SQP} algorithm for
  large-scale constrained optimization.
\newblock \emph{SIAM Review (SIREV)} 47: 99--131.

\bibitem[{Inga et~al.(2019)Inga, Bischoff, K{\"o}pf and
  Hohmann}]{inga2019arxiv}
Inga J, Bischoff E, K{\"o}pf F and Hohmann S (2019) Inverse dynamic games based
  on maximum entropy inverse reinforcement learning.
\newblock \emph{arXiv preprint arXiv:1911.07503} .

\bibitem[{Isaacs(1954-1955)}]{isaacs1954differential}
Isaacs R (1954-1955) Differential games i-iv.
\newblock Technical report, RAND CORP SANTA MONICA CA SANTA MONICA.

\bibitem[{Jin et~al.(2021)Jin, Kuli{\'c}, Mou and Hirche}]{jin2021inverse}
Jin W, Kuli{\'c} D, Mou S and Hirche S (2021) Inverse optimal control from
  incomplete trajectory observations.
\newblock \emph{Intl.~Journal~of Robotics Research (IJRR)} 40(6-7): 848--865.

\bibitem[{Kalman(1964)}]{kalman1964jbe}
Kalman RE (1964) {When Is a Linear Control System Optimal?}
\newblock \emph{ASME Journal of Basic Engineering} 86(1): 51--60.
\newblock \doi{10.1115/1.3653115}.

\bibitem[{Keshavarz et~al.(2011)Keshavarz, Wang and Boyd}]{keshavarz2011isic}
Keshavarz A, Wang Y and Boyd S (2011) Imputing a convex objective function.
\newblock In: \emph{Proc.~of the Intl.~Symp. on Intelligent Control (ISIC)}.
  IEEE.

\bibitem[{K{\"o}pf et~al.(2017)K{\"o}pf, Inga, Rothfu{\ss}, Flad and
  Hohmann}]{kopf2017ifac}
K{\"o}pf F, Inga J, Rothfu{\ss} S, Flad M and Hohmann S (2017) Inverse
  reinforcement learning for identification in linear-quadratic dynamic games.
\newblock \emph{IFAC-PapersOnLine} 50(1): 14902--14908.

\bibitem[{Kretzschmar et~al.(2016)Kretzschmar, Spies, Sprunk and
  Burgard}]{kretzschmar2016socially}
Kretzschmar H, Spies M, Sprunk C and Burgard W (2016) Socially compliant mobile
  robot navigation via inverse reinforcement learning.
\newblock \emph{Intl.~Journal~of Robotics Research (IJRR)} 35(11): 1289--1307.

\bibitem[{Laine et~al.(2021)Laine, Fridovich-Keil, Chiu and
  Tomlin}]{laine2021arxiv}
Laine F, Fridovich-Keil D, Chiu CY and Tomlin C (2021) The computation of
  approximate generalized feedback nash equilibria.
\newblock \emph{arXiv preprint arXiv:2101.02900} .

\bibitem[{Le~Cleac'h et~al.(2020)Le~Cleac'h, Schwager and
  Manchester}]{cleac2020rss}
Le~Cleac'h S, Schwager M and Manchester Z (2020) {ALGAMES}: A fast solver for
  constrained dynamic games.
\newblock In: \emph{Proc.~of Robotics: Science and Systems (RSS)}.

\bibitem[{Le~Cleac'h et~al.(2021)Le~Cleac'h, Schwager and
  Manchester}]{cleac2020arxiv}
Le~Cleac'h S, Schwager M and Manchester Z (2021) {LUCIDGames}: Online unscented
  inverse dynamic games for adaptive trajectory prediction and planning.
\newblock \emph{IEEE Robotics and Automation Letters (RA-L)} 6(3): 5485--5492.

\bibitem[{Levine and Koltun(2012)}]{levine2012icml}
Levine S and Koltun V (2012) Continuous inverse optimal control with locally
  optimal examples.
\newblock \emph{Proc.~of the Int.~Conf.~on Machine Learning (ICML)} .

\bibitem[{Luo et~al.(1996)Luo, Pang and Ralph}]{luo1996mathematical}
Luo ZQ, Pang JS and Ralph D (1996) \emph{Mathematical programs with equilibrium
  constraints}.
\newblock Cambridge University Press.

\bibitem[{Menner and Zeilinger(2020)}]{menner2020arxiv}
Menner M and Zeilinger MN (2020) Maximum likelihood methods for inverse
  learning of optimal controllers.
\newblock \emph{arXiv preprint arXiv:2005.02767} .

\bibitem[{Mombaur et~al.(2010)Mombaur, Truong and Laumond}]{mombaur2010ar}
Mombaur K, Truong A and Laumond JP (2010) From human to humanoid
  locomotion—an inverse optimal control approach.
\newblock \emph{Autonomous Robots} 28(3): 369--383.

\bibitem[{Monderer and Shapley(1996)}]{monderer1996potential}
Monderer D and Shapley LS (1996) Potential games.
\newblock \emph{Games and economic behavior} 14(1): 124--143.

\bibitem[{Mukadam et~al.(2019)Mukadam, Dong, Dellaert and
  Boots}]{mukadam2019steap}
Mukadam M, Dong J, Dellaert F and Boots B (2019) Steap: simultaneous trajectory
  estimation and planning.
\newblock \emph{Autonomous Robots} 43(2): 415--434.

\bibitem[{Natarajan et~al.(2010)Natarajan, Kunapuli, Judah, Tadepalli, Kersting
  and Shavlik}]{natarajan2010multi}
Natarajan S, Kunapuli G, Judah K, Tadepalli P, Kersting K and Shavlik J (2010)
  Multi-agent inverse reinforcement learning.
\newblock In: \emph{2010 ninth international conference on machine learning and
  applications}. IEEE, pp. 395--400.

\bibitem[{Ng and Russell(2000)}]{ng2000icml}
Ng AY and Russell SJ (2000) Algorithms for inverse reinforcement learning.
\newblock In: \emph{Proc.~of the Int.~Conf.~on Machine Learning (ICML)}.

\bibitem[{Nocedal and Wright(2006)}]{nocedal2006optimizationbook}
Nocedal J and Wright S (2006) \emph{Numerical optimization}.
\newblock Springer Verlag.

\bibitem[{Peters(2020)}]{peters2020master}
Peters L (2020) \emph{Accommodating Intention Uncertainty in General-Sum Games
  for Human-Robot Interaction}.
\newblock Master's Thesis, {Hamburg University of Technology}.

\bibitem[{Peters et~al.(2021)Peters, Fridovich-Keil, Rubies-Royo, Tomlin and
  Stachniss}]{peters2021rss}
Peters L, Fridovich-Keil D, Rubies-Royo V, Tomlin CJ and Stachniss C (2021)
  Inferring objectives in continuous dynamic games from noise-corrupted partial
  state observations.
\newblock In: \emph{Proc.~of Robotics: Science and Systems (RSS)}.

\bibitem[{Rothfu{\ss} et~al.(2017)Rothfu{\ss}, Inga, K{\"o}pf, Flad and
  Hohmann}]{rothfuss2017ifac}
Rothfu{\ss} S, Inga J, K{\"o}pf F, Flad M and Hohmann S (2017) Inverse optimal
  control for identification in non-cooperative differential games.
\newblock \emph{IFAC-PapersOnLine} 50(1): 14909--14915.

\bibitem[{{\v{S}}o{\v{s}}i{\'c} et~al.(2016){\v{S}}o{\v{s}}i{\'c}, KhudaBukhsh,
  Zoubir and Koeppl}]{vsovsic2016inverse}
{\v{S}}o{\v{s}}i{\'c} A, KhudaBukhsh WR, Zoubir AM and Koeppl H (2016) Inverse
  reinforcement learning in swarm systems.
\newblock \emph{arXiv preprint arXiv:1602.05450} .

\bibitem[{W{\"a}chter and Biegler(2006)}]{wachter2006jmp}
W{\"a}chter A and Biegler LT (2006) On the implementation of an interior-point
  filter line-search algorithm for large-scale nonlinear programming.
\newblock \emph{Mathematical Programming} 106(1): 25--57.

\bibitem[{Wang et~al.(2019)Wang, Spica and Schwager}]{wang2019dars}
Wang Z, Spica R and Schwager M (2019) Game theoretic motion planning for
  multi-robot racing.
\newblock \emph{Distributed Autonomous Robotic Systems} : 225--238.

\bibitem[{Ziebart et~al.(2008)Ziebart, Maas, Bagnell and Dey}]{ziebart2008aaai}
Ziebart BD, Maas AL, Bagnell JA and Dey AK (2008) Maximum entropy inverse
  reinforcement learning.
\newblock In: \emph{Proc.~of the Conference on Advancements of Artificial
  Intelligence (AAAI)}.

\end{thebibliography}

\end{document}